\documentclass[10pt,twocolumn,letterpaper]{article}

\usepackage[pagenumbers]{cvpr} 

\usepackage{graphicx}
\usepackage{amsmath}
\usepackage{amssymb}
\usepackage{booktabs}
\usepackage{multirow}
\usepackage{hhline}
\usepackage{bbding} 
\usepackage{hyperref}

\usepackage[]{hyperref}

\usepackage[capitalize]{cleveref}
\crefname{section}{Sec.}{Secs.}
\Crefname{section}{Section}{Sections}
\Crefname{table}{Table}{Tables}
\crefname{table}{Tab.}{Tabs.}

\def\cvprPaperID{378} 
\def\confName{CVPR}
\def\confYear{2022}

\begin{document}

\title{FERV39k: A Large-Scale Multi-Scene Dataset for Facial Expression\\  Recognition in Videos}


\author{Yan Wang$^{1}$, Yixuan Sun$^{1}$, Yiwen Huang$^{2}$, Zhongying Liu$^{2}$, Shuyong Gao$^{2}$,\\ Wei Zhang$^{2}$, Weifeng Ge$^{2,*}$ and Wenqiang Zhang$^{1,2,*}$ \\
 $^{1}$Academy of Engineering \& Technology, Fudan University, Shanghai, China\\
 $^{2}$School of Computer Science, Fudan University, Shanghai, China\\
{\tt\small \{wfge and wqzhang\}@fudan.edu.cn}
}

\twocolumn[{
\vspace{-0.3cm}
\maketitle 
\vspace{-0.6cm}
\renewcommand\twocolumn[1][]{#1}%
\vspace{-0.4cm}
\begin{center}
\centering
\includegraphics[width=\textwidth]{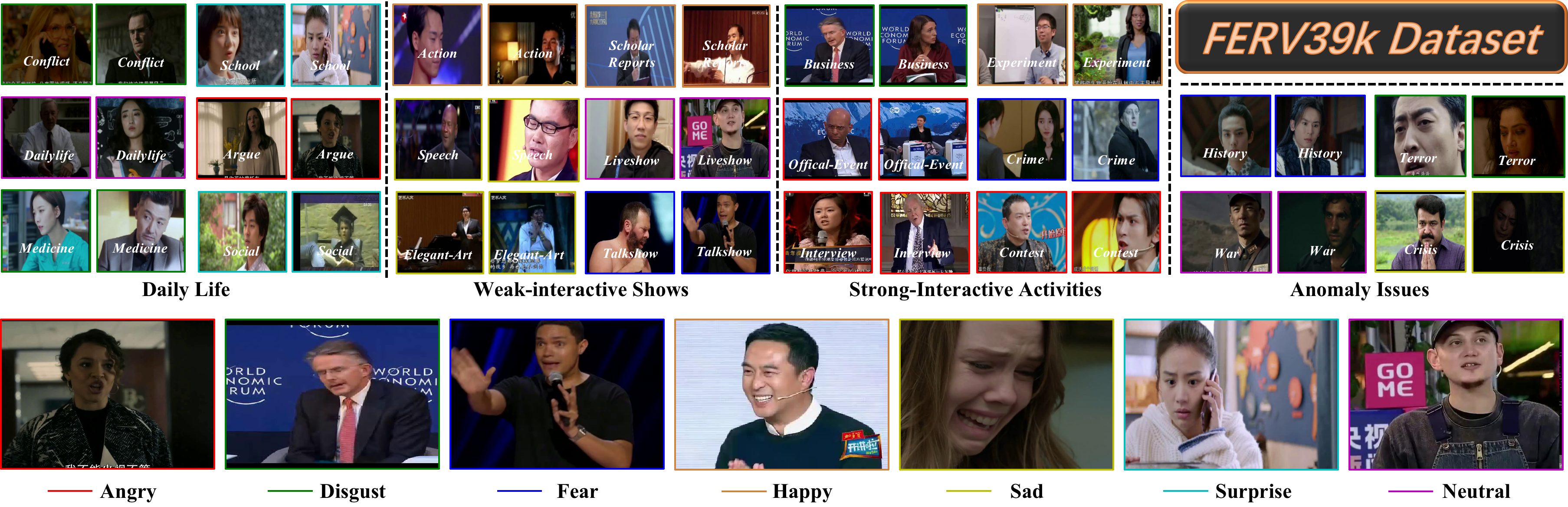}
\vspace{-0.7cm}
\captionof{figure}{An overview of FERV39k composed by video frames of 7 basic expression across 4 scenarios subdivided by 22 scenes.}\label{fig:fig1}
\vspace{-0.1cm}\end{center}}]

\let\thefootnote\relax\footnotetext{$^*$ Corresponding author}

\begin{abstract}
\vspace{-0.2cm}
    Current benchmarks for facial expression recognition (FER) mainly focus on static images, while there are limited datasets for FER in videos. It is still ambiguous to evaluate whether performances of existing methods remain satisfactory in real-world application-oriented scenes. For example, “Happy” expression with high intensity in Talk-Show is more discriminating than the same expression with low intensity in Official-Event. To fill this gap, we build a large-scale multi-scene dataset, coined as FERV39k. We analyze the important ingredients of constructing such a novel dataset in three aspects: (1) multi-scene hierarchy and expression class, (2) generation of candidate video clips, (3) trusted manual labelling process. Based on these guidelines, we select 4 scenarios subdivided into 22 scenes, annotate 86k samples automatically obtained from 4k videos based on the well-designed workflow, and finally build 38,935 video clips labeled with 7 classic expressions. Experiment benchmarks on four kinds of baseline frameworks were also provided and further analysis on their performance across different scenes and some challenges for future research were given. Besides, we systematically investigate key components of DFER by ablation studies. The baseline framework and our project will be available.
    
    
\end{abstract}

\vspace{-0.5cm}


\begin{table*}[ht]
\centering
\footnotesize
\vspace{-0.2cm}
\setlength\tabcolsep{0.014\linewidth}
\begin{tabular}{lccccccccccc}
\toprule
\multirow{2}{*}{Database (Year)} & \multirow{2}{*}{Samples} & \multirow{2}{*}{Emo.} & \multirow{2}{*}{Anno.} & \multirow{2}{*}{Best.} & \multirow{2}{*}{Context} & \multirow{2}{*}{Scene} & \multicolumn{5}{c}{Video Sources}               \\ \cline{8-12} 
                                 &                          &                         &                             &                           &                          &                        & Lab Shot & Movie & TV & Live Show & Others \\ \midrule
CK+ (2010) \cite{W13_lucey2010extended}                      & 327                      & 7 Exps                 & 1                           & 99.69                     & Lab                      &                        & \textbf{$\checkmark$}        &       &         &           &        \\
Oulu-CASIA (2011) \cite{W14_zhao2011facial}               & 560                      & 6 Exps                 & 1                           & 92.7                     & Lab                      &                        & \textbf{$\checkmark$}        &       &         &           &        \\
Aff-Wild (2017) \cite{W24_kollias2019deep}                 & 298                      & V-A                     & 8                           & N/A                       & Wild                     &                        &          & \textbf{$\checkmark$}     & \textbf{$\checkmark$}       &           & \textbf{$\checkmark$}     \\
AFEW-VA (2017)  \cite{W18_kossaifi2017afew}                 & 600                      & V-A                     & 2                           & N/A                       & Wild                     &                        &          & \textbf{$\checkmark$}     &         &           &        \\
AFEW 8.0 (2018) \cite{W11_dhall2018emotiw}                 & 1,809                     & 7 Exps                 & 2                           & 53.26                     & Wild                     &                        &          & \textbf{$\checkmark$}     &         &           &        \\
CAER  (2019)  \cite{W19_lee2019context}                   & 13,201                   & 7 Exps                 & 3                           & 77.04                     & Wild                     &                        &          &       & \textbf{$\checkmark$}       &           &        \\
DFEW (2020) \cite{W12_jiang2020dfew}                     & 16,372                   & 7 Exps                 & 10                          & 56.41                     & Wild                     &                        &          & \textbf{$\checkmark$}     & \textbf{$\checkmark$}        &           &       \\
FERV39k (2021)                      & 39,546                   & 7 Exps                 & 30                      & N/A                   & Wild                     & \textbf{$\checkmark$}                      &          & \textbf{$\checkmark$}     & \textbf{$\checkmark$}       & \textbf{$\checkmark$}         & \textbf{$\checkmark$}      \\ \bottomrule
    \end{tabular}
    \vspace{-0.3cm}
    \caption{Comparison of statistics of existing available DFER datasets and our built FERV39k. (Emo. = Emotion distribution; Anno. = Annotation times; Best. =Best accuracy; Exps = Expressions; V-A = Valence-Arousal.)}
    \label{tab:tab1}
    \vspace{-0.6cm}
\end{table*}

\section{Introduction}
\label{sec:intro}

Facial expression recognition (FER) in static images \cite{W1_wang2020region} or videos \cite{W2_liu2020saanet} is of great importance to many applications, such as human-computer interaction (HCI) \cite{W3_azazi2015towards} and lie detection \cite{W5_barman2019facial}. With millions of images uploaded every day by users from different events and social gatherings, there are various available large-scale datasets for static FER, such as RAF-DB \cite{W6_li2017reliable} and AffectNet \cite{W7_mollahosseini2017affectnet}. On top of these datasets, various methods \cite{W8_farzaneh2021facial, W9_fu2020semantic, W10_li2020deep} are designed to understand human emotion and recognize facial expressions. In contrast to static image FER, there are only a few video-based facial expression datasets. In early period, researchers paid attention to in-the-lab datasets, such as CK+ \cite{W13_lucey2010extended} and Oulu-CASIA \cite{W14_zhao2011facial}, which are collected from lab environment and contain limited posed video clips with no more than 30 frames. Recently, recognizing expressions from in-the-lab short video clips has achieved considerable progress \cite{W15_zhao2016peak, W16_yu2018spatio, W17_meng2019frame, W2_liu2020saanet}, but these models often fail to be directly applied for in-the-wild scenes. Typically, limited samples without complex and varied scene context might be impractical for real-world applications.

With the development of AFEW competition \cite{W11_dhall2018emotiw, W18_kossaifi2017afew}, video-based in-the-wild datasets are released progressively, but their video clips are limited and not enough for developing deep FER models. Although the seeming datasets, such as CAER \cite{W19_lee2019context} and DEFW \cite{W12_jiang2020dfew}, claim that their sources of videos are diverse, there exist some limitations in these datasets. For CAER \cite{W19_lee2019context}, data volume reaches 13k, however, its scene is single and lacks of challenge to FER methods. DEFW \cite{W12_jiang2020dfew} is a large-scale and well-annotated unconstrained dataset for FER in videos, but it fails to consider and further differentiate scene categories \cite{W20_cowen2021sixteen}, which are essential for application-oriented expression recognition. Besides, these works all overlook how to automatically generate abundant candidate video clips for manual annotation to meet the need of building a larger-scale dataset.

It is necessary to build a multi-scene dataset to advance the FER in video. The benchmark should satisfy several important requirements to cover realistic challenges. 1) Considering the complexity of real-world applications, selected scenes should cover various aspects. 2) With billions of videos currently accessed from Internet and video platforms, there is an urgent need for robust algorithms that can automatically generate the massive video clips. 3) Due to the complexity of facial expression annotations, the workflow of annotating video clips needs to be well-designed. Based on above guidelines, we build FERV39k (Figure \ref{fig:fig1}), which is a large-scale, multi-scene, high-quality dataset, and contains 38,935 video clips labeled with 7 classic expressions in 4 scenarios: Daily Life, Weak-Interactive Shows, Strong-Interactive Activities, and Anomaly Issues.We design scenarios and scenes following four reasons: 1) Plenty of video sources and samples. 2) Expandability of 22 fine-grained scenes. 3) Large variations and limited overlapping. 4) Distinct associations with scene context. Besides, we design a four-stage strategy, which itself generates 86k candidate video clips from 4k raw videos. 

Specifically, our built FERV39k has 3 main characteristics: 1) Multi-scene: clips are divided into 4 scenarios and subdivided into 22 scenes with different characteristics. 2) Large-scale: the amount of video clips reaches 39k with last time from 0.5s to 4s, which indicates that available video frames and cropped facial images reach 1M with resolution of $336\times504$, and $224\times224$, respectively. 3) High-quality: workflow of crowdsourcing and professional annotation is adopted to ensure high-quality labels with guidance of fine-grained expressions.

Given the well-annotated and multi-scene video clips in our built dataset, we first benchmark four kinds of deep learning-based architectures for FER in videos on the challenging FERV39k following action recognition baselines \cite{W21_kay2017kinetics, liu2021no, W22_carreira2017quo}. We then perform several baseline evaluations with four baselines and representative backbones to reveal challenging aspects of multi-scene expression representation in videos. According to our analysis on FERV39k benchmark, we uncover several new challenges: 1) difficulty and confusion of 7 basic expression classes. 2) discrepancy across 4 scenarios. 3) unsatisfactory cross-scenario performance. 4) long-tail distribution of expressions and duration. In order to systematically enumerate key components in modeling DFER based on the four baseline architectures on FERV39k, we further carry out several ablation studies and figure out some significant findings: 1) Pre-training on large-scale datasets is not always helpful. 2) More sampling fails to steadily improve the performance. 3) Scene information plays a complementary role on DFER.

In summary, our work has three main contributions: 1) We construct a novel large-scale multi-scene FERV39k dataset for both intra-scene and inter-scene DFER. The dataset contains 38,935 video clips labeled with 7 classic expressions across 22 fine-grained scenes in 4 isolated scenarios. To our best knowledge, this is the first dynamic FER dataset with 39K clips, scenario-scene division as well as the cross-domain supportability. 2) We proposed four-stage candidate clip generation and two-stage annotation workflow with a balance between cost and quality control which can be used in other large-scale facial video dataset construction. 3) We benchmark four kinds of deep learning-based architectures and conduct in-depth studies of FERV39k, which reveal the key challenges of our dataset and indicate new directions of future research according to extensive ablation studies.



\section{Related Work}
\label{sec:Related_Work}

\subsection{Video-based Datasets for DFER}

Video-based FER datasets \cite{W14_zhao2011facial, W11_dhall2018emotiw, W19_lee2019context} have been proposed since the start of the researches on facial expressions. In earlier time, the participants were required or induced to perform targeted facial expressions in the controlled environments to collect data such as CK+ \cite{W13_lucey2010extended} and Oulu-CASIA \cite{W14_zhao2011facial}. However, subject to the scale of participants and experiment conditions, in-the-lab datasets are usually small-scaled and in which facial expressions are usually far from the real-world expressions. Besides, most methods \cite{otberdout2020dynamic,W2_liu2020saanet} (Table \ref{tab:tab1}) have already obtained excellent performance on these benchmarks. 

As a result, more attentions are attracted by datasets collected from in-the-wild conditions with naturalistic emotion states, such as AFEW \cite{W11_dhall2018emotiw}, Aff-Wild \cite{W24_kollias2019deep}, AFEW-VA \cite{W18_kossaifi2017afew}, CAER \cite{W19_lee2019context}, and DFEW \cite{W12_jiang2020dfew}. AFEW \cite{W11_dhall2018emotiw} is the first in-the-wild dataset proposed in 2013 which contains 1,809 clips of 330 subjects labeled for twice with seven labels. AFEW-VA \cite{W18_kossaifi2017afew} provides more subjects, samples, and professional annotations as well as valence-arousal annotation. CAER \cite{W19_lee2019context} increases the number of video clips to 13,201, and considers cropped face and context information. DFEW \cite{W12_jiang2020dfew} expands the scale and diversity of data and improves the annotation quality. Table \ref{tab:tab1} compares statistics among existing datasets with our built FERV39k, which has the following characteristics: 1) the largest number of samples reaches 39k obtained from 86k automatically generated candidate video clips. 2) the well-designed workflow of annotation in combination of crowdsourcing and professional review. 3) The hierarchy design of two-level scenes is creative to help application-oriented DFER and Cross-domain learning in different contexts. 4) All raw videos are collected from cross-platform sources. 

\subsection{Dynamic FER Approaches}

Though various methods can recognize expressions from static images \cite{W10_li2020deep}, the dynamic videos usually contain more information including the movement of appearance as well as other temporal information. There are two kinds of network structures, named 3-dimensional convolutional networks (3D ConvNet) and 2D ConvNet-LSTM, commonly used for DFER. The 3D ConvNet-based methods \cite{W26_tran2015learning, W28_lo2020mer} use 3D ConvNet extracting spatio-temporal features and generating embedding for DFER. For example, the works \cite{W26_tran2015learning,W27_al2018deep} use C3D \cite{W26_tran2015learning} for local spatio-temporal feature extraction. The 2D ConvNet-LSTM based methods \cite{zhang2017facial, W32_ma2019emotion, W2_liu2020saanet} combine the CNNs and the LSTM for extracting spatial features and learning temporal modeling, respectively. Most works \cite{W31_fan2016video} mainly rely on analysis of cropped face regions, ignoring scene context information for emotion recognition in the wild. To solve these limitations, Lee et al. \cite{W19_lee2019context} investigated the influence of context information by two-stream encoding network (CAER-Net) which consists of face encoding stream and context encoding stream, to separately encode the cropped face region and context information. With analysis and comparison among existing video-based representation architectures on whether convolutional layers use 2D or 3D kernels, and whether the input to network includes scene context, we design four kinds of baseline architectures.


\section{The FERV39k Dataset}
\label{sec:MULTI_SCENE}
In order to introduce a novel and challenging benchmark for application-oriented DFER, we propose a well-designed procedure of dataset construction to build our FERV39k with high-quality annotations. The FERV39k is more challenging and inspiring than previous ones in multiple application scenes, cross-domain learning supportability, automatic candidate clip selection and two-stage efficient \& highly credible annotation. While other types of annotations based on these data will be included in succeeding versions, e.g., frame-level annotation with key expression, the current version of FERV39k mainly provides annotations for DFER on 4 isolated scenarios with 22 fine-grained scenes labeled by 7 basic expressions.

\subsection{Key Challenges}

Inspired by the key challenges \cite{W34_shao2020finegym}, we consider a series of unprecedented difficulties and scheme the corresponding strategies, which are followed as: 

\noindent{\textbf{How to define and generate the scenes and expressions?}} Since thousands of contexts/scenes and dozens of facial expressions occurred systematically in all countries, it is impractical to fulfill the all-scene task in work\cite{W20_cowen2021sixteen}. Fortunately, we analyze the findings and conclusions from the work of Cowen et al. \cite{W20_cowen2021sixteen}, which help us summarize 4 scenarios consisting of 22 scenes as well as the 7 basic expressions. Furthermore, novel scene-based keyword list and fine-grained labels are designed. 

\noindent{\textbf{How to automatically generate candidate video clips?}} Different from static facial images crawled from the Internet based on keywords \cite{W7_mollahosseini2017affectnet}, extra segmentation is required to obtain short-duration video clips with single expression due to the story complexity of a video or movie. Generally, the pipeline of candidate video clips collection for a DFER dataset is crawling large-scale videos (metadata) from the Internet and cropping the single expression clips manually. However, manual operation is costly for a large-scale dataset. Therefore, a novel four-stage FER-based video segmentation process is proposed. 

\begin{figure}[!t]
\vspace{-0.2cm}
  \centering
  \includegraphics[width=1.0\linewidth]{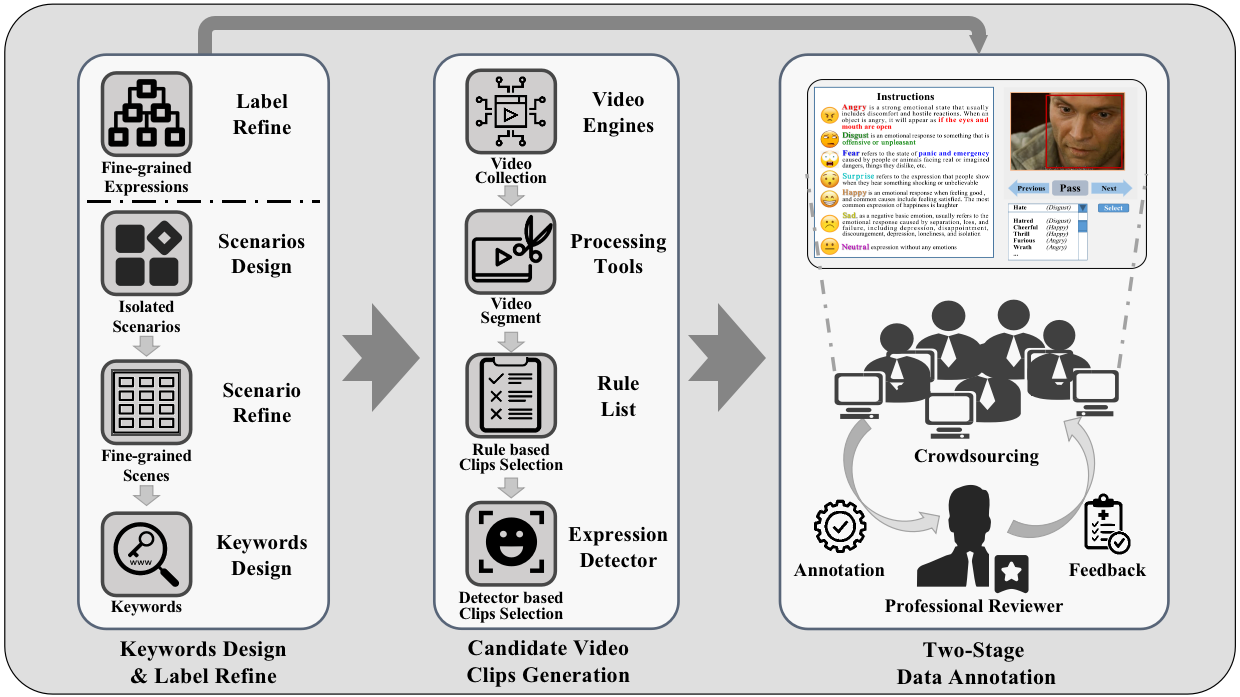}
  \vspace{-0.6cm}
  \caption{An overview of FERV39k construction.}
  \label{fig:fig2}
  \vspace{-0.8cm}
\end{figure}

\noindent{\textbf{How to design annotation procedure with quality control?}} Crowdsourcing services such as Amazon Mechanical Turk or JD Crowdsourcing are commonly used to build a large-scale dataset. However, discovering subtle difference between some expressions requires professional knowledge. As a result, a two-stage annotation workflow is proposed to get quality-guaranteed annotation with a balance between cost and reliability. 


\subsection{Dataset Construction Procedure}

In Sec. 3.2, we will introduce the three steps for dataset construction named selection  of  scene  vocabulary  and  expression  class, generation  of  candidate  video  clips and data annotation (Figure \ref{fig:fig2}). 

\begin{figure*}[ht]
\vspace{-0.1cm}
  \centering
   \includegraphics[width=1.0\linewidth]{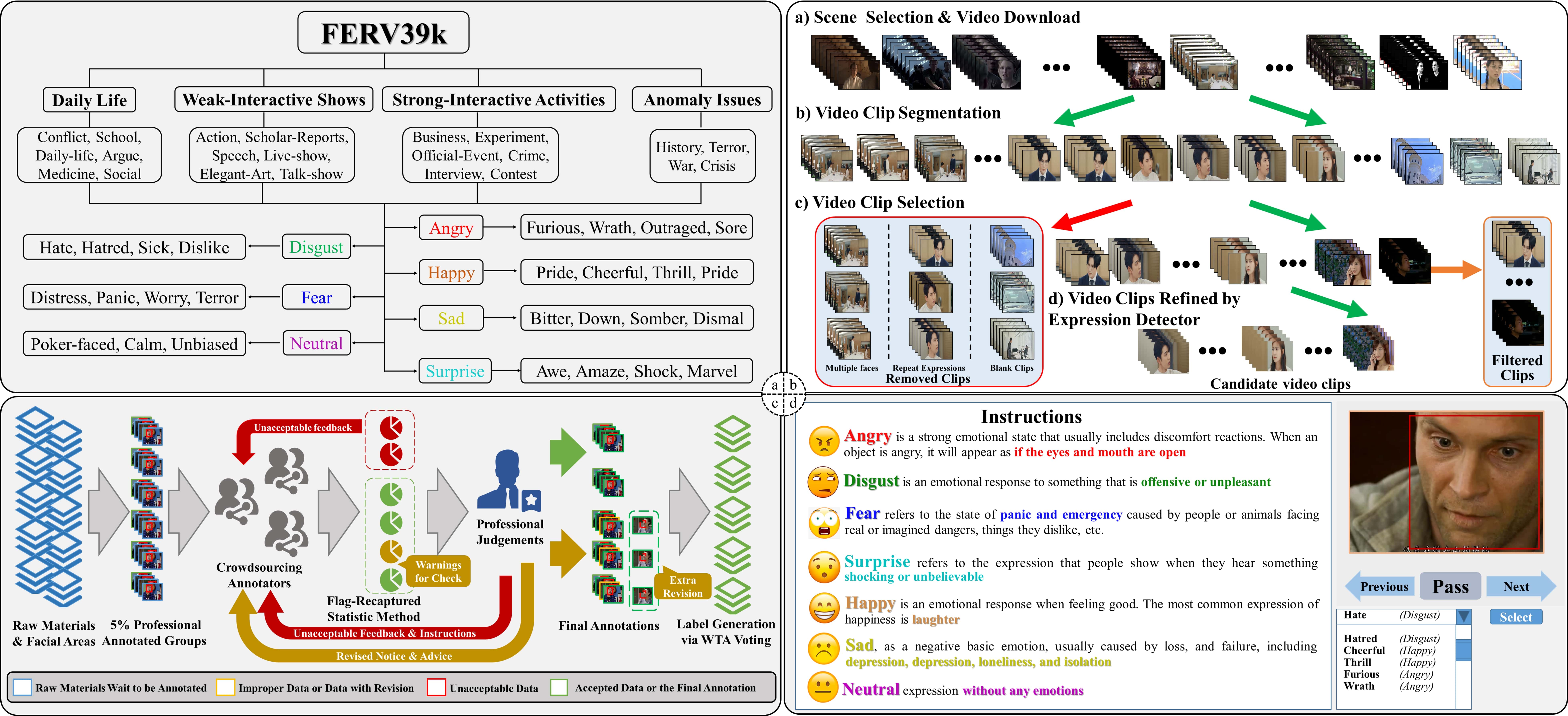}
   \vspace{-0.6cm}
   \caption{Four important components for dataset construction. (a) Our design of 4 isolated scenarios, 22 scenes, 7 basic expressions and 26 fine-grained expressions. (b) The four-stage generation of candidate video clips in the FERV39k dataset. (c) The procedure of data annotation, statistic evaluation, professional judgement and label generation. (d) The labelling interface used in crowd sourcing platform.}
   \label{fig:fig3}
   \vspace{-0.5cm}
\end{figure*}

\noindent{\textbf{Selection of scene vocabulary and expression class.}}  Before data collection, we first design the Scene Vocabulary (including their keywords) and Expression Class in parallel. For the Scene Vocabulary generating, we analyze the statistic result from work \cite{W20_cowen2021sixteen}, select 22 representative scenes and divide them into 4 scenarios: 6 scenes \{Argue, Social, School, Medicine, Conflict, and Daily-Life\} designed for Daily Life (DL11k), 6 scenes \{Action, Scholar-Reports, Speech, Elegant-Art, Live-Show, and Talk-Show\} designed for Weak-Interactive Shows (WIS9k), 6 scenes \{Business, Experiment, Official-Event, Crime, Interview, Contest\} for Strong-Interactive Activities (SIA10k) and 4 scenes \{History, Terror, War and Crisis\} for Anomaly Issues (AI9k). For our scene-based raw material collection, we also design a keyword list for each scene. And to design our Expression Class, 7 basic expressions namely “Angry”, “Disgust”, “Fear”, “Happy”, “Sad”, “Surprise”, “Neutral” are selected as annotation labels. And we follow the taxonomy defined by Parrot etc. \cite{W35_volkova2014emotion} to carefully select 26 words \cite{W37_liang2020fine} aiming at clarifying the difference of fine-grained emotion classes, which results in the final expression hierarchy shown in Figure \ref{fig:fig3}(a). Following to expression definition \cite{W7_mollahosseini2017affectnet, W11_dhall2018emotiw, W12_jiang2020dfew}, we also initialize an expression list and write a handbook to clarify each expressions. 

%
%
%
%
%
%
%
%
%
%
%
%
%
%
%
%
%
%
%
%
%

\noindent{\textbf{Generation of candidate video clips.}} Following the works \cite{W6_li2017reliable, W7_mollahosseini2017affectnet, W24_kollias2019deep, W12_jiang2020dfew}, online videos originate from the real-life environments in different scenes, hence the human expression in the videos can be recognized as real-world facial expression. We start with reviewing top-level 22 scenes and then collecting corresponding online videos, TV shows and movies from searching/video engines. To acquire clips, existing works ask annotators to manually segment video clips with expressions via video editing software. For processing data in smaller scale, the cost of time and labor is affordable. However, for our 39k clips dataset (raw materials are even more), it seems impractical to extract clips manually. Hence, we adopt a four-stage strategy to collect and generate candidate video clips for multi-scene videos, the pipeline of which is shown in Figure \ref{fig:fig3}(b).

Firstly, we download over 6k meta data with different lasting time from 8 world-wide open-source engines containing Asian, African, and European/American videos via generated keyword list. Afterwards we sort and randomly remove some of the videos. After this step, 4k pieces of data with balanced time distribution of scenes left. According to work \cite{W36_ben2021video}, we randomly segment them into video clips among 0.5-4 seconds. In order to generate facial clips, we make a rule list to help our well-designed mechanism to adaptively and automatically select a twenty-fold amount of clips than expected scale of final dataset. However, the rule-based selection mechanism is rough for generating a good candidate and manual refinement is still a hard work. As a result, we utilize a pretrained light-weight ResNet-50 FER detector to refine these clips and generate candidate clips with expression predictions. Finally, with a prospect that the scale of filtered clips is double of final dataset scale, we randomly remove some clips and keep the latency distribution of estimated expressions fit to the real-world (work \cite{W20_cowen2021sixteen}). 

\noindent{\textbf{Manual annotation.}} To achieve the balance of professional annotation and cost control, we design a workflow of annotation-examine for data annotation (Figure \ref{fig:fig3}(c)). In our designed procedure, there are two roles named crowd-sourcing annotator (CA, 20 workers) and professional researcher (PR, 10 workers), respectively. Our goal is subtly using PRs to get professional annotation at lower cost. To further help annotators differentiate our task from many others on the platform as well as make our task as stimulating and engaging as possible, the JD Crowdsourcing establishes a single-page web base on our guidance. The labelling interface is shown in Figure \ref{fig:fig3}(d), in which one video clip, introductions and the bounding box of face area in each frame are provided to assist annotators. Besides, the platform can automatically convert 26 choices into 7 expression labels.

The clips are divided into groups at first (5\% of each are PR annotated) and copied 3 times. Then we randomly shuffle the grouped materials and provide them to CAs. CAs are asked to choose the most likely word or “PASS” on the platform. After annotation, group copies are checked via Flag-Recaptured Statistic method \cite{bell1974population}. We design 80\% and 40\% correct rate as two thresholds and mark copies as unacceptable (UA), Improper (IP) and Accept (AC). The IP and AC groups will be passed to PRs for judgement. In this step PRs only need to decide whether the annotation of a group is acceptable. The UA ones will be retreated to CAs and ones still IP will be relabeled by the PRs. For both UA and IP, PRs will provide feedback to CAs. Afterwards, the Weighted-Winner-take-all (WWTA) voting method is used for generating final facial expression labels. To our goal, after iterations on a few groups, the annotators can provide relatively reliable annotations and verification work will become less complex. 


\subsection{Dataset Statistics}
\begin{figure*}[!ht]
  \centering
   \includegraphics[width=1.0\linewidth]{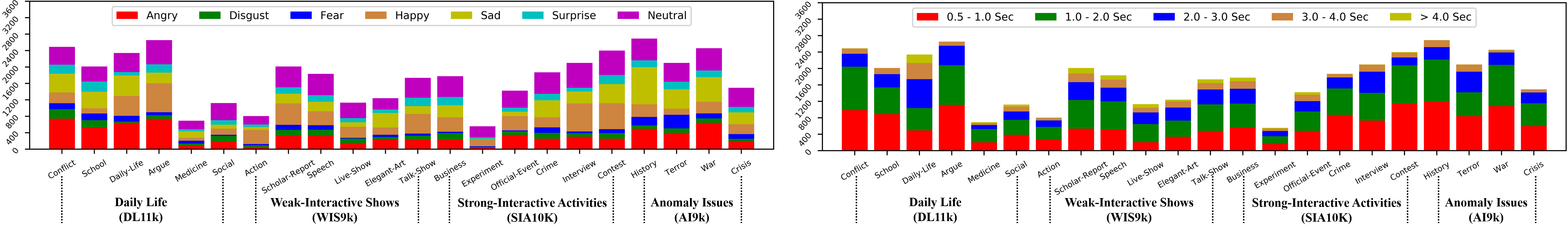}
   \vspace{-0.6cm}
   \caption{Statistics of our FERV39k. Left chart is the distribution of 7 expressions of video clips in different scenes, sorted by 4 scenarios and 7 different expressions. And Right chart is the distribution of 7 expression duration of video clips in different scenes, which is sorted by 4 scenarios and 5 different time duration.}
   \label{fig:fig5}
    \vspace{-0.5cm}
\end{figure*}

The FERV39k consists of 4 isolated scenarios subdivided by 22 detailed scenes, including nearly 39k video clips labeled with 7 basic facial expressions \cite{W10_li2020deep} with average duration for 1.5 seconds. In general, the clips are evenly distributed in 4 scenarios but the scale of each scene also reflects a severe long-tailed distribution. For further analysis, Figure \ref{fig:fig5} (left) shows the number of clips in each scene and the distribution of expressions in our built FERV39k, which is used for baseline analysis in this paper. The histogram chart shows a natural long-tailed distribution across 7 basic expressions in different scenes. For instance, ‘Fear’ appears more in the ‘Terror’ scene (18\%) and ‘Happy’ appears more in the ‘Live-Show’ scene (33\%). This will be a new challenge for DFER models. Figure \ref{fig:fig5} (right) shows the distribution of expression duration of video clips in different scenes. The large variation of expression duration makes it more difficult for DFER models to accurately localize key frames like \cite{W34_shao2020finegym}. Moreover, expression instances in FERV39k are often related with longer temporal context and interactions with context. These inherent challenges of FERV39k require a more powerful and flexible temporal modeling scheme for expression detection. \textbf{Our built FERV39k can be available under the condition of abiding by the agreement.} 

\subsection{Dataset Characteristics}

Our FERV39k has several distinguishing and attracting characteristics compared with existing datasets. 

\noindent{\textbf{Large-scale candidate video clips.}} With the introduction of four-stage candidate clip generation method, we can cheaply acquire massive candidate video clips, which makes FERV39k possible to be further expanded.

\noindent{\textbf{High-quality annotation.}} With our two-stage annotation strategy, supporting files, fine-grained choices as well as Flag-Recaptured Statistic methods, Professional Judgement and WWTA Voting, FERV39k can get reliable labels at lower cost.

\noindent{\textbf{Task difficulty.}} With 4 difficulties proposed: 1) large variance of expression duration among clips; 2) different intensities of expressions across different scenes; 3) limited representing frames for labeled expression in a clip; 4) severe long-tailed distribution in different scenes and expressions, FERV39k brings new challenges for DFER methods.

\noindent{\textbf{Application-oriented diversity.}} With a new sight of application, FERV39k pays attention to specific application performance and cross-scene robustness of DFER methods.

\section{Benchmark Performance}
\label{sec:Experiments}

In this section, we will conduct experiments to show the challenges of FERV39k in practical via baseline evaluations and figure out some findings via ablation studies.

\subsection{Experiment Setup}

\noindent{\textbf{FERV39k protocol.}} To build a solid DFER benchmark, we manually split all data into training set (including validation set), and testing set. In FERV39k Benchmark, video clips of all scenes are randomly shuffled and split into training (80\%), and testing (20\%) without overlapping, which forms 27 kinds of configurations consisting of 22 setups for each scene, 4 setups for isolated scenarios and 1 setup for all scenes. Cross-scene learning is also available for which some special scenes are used for testing. Besides, we provide cropped face images with $224\times224$ resolution and scene images with $336\times504$ resolution to meet the requirement of context-aware DFER methods.

\noindent{\textbf{Implementation details.}} In our experiments, the whole framework is built on PyTorch-GPU using NVIDIA GeForce RTX 2080Ti GPUs. We set learning rate (lr) in a range between 1e-3 and 1e-2, weight decay as 1e-4, and the batch size is fixed at 32 for all architectures. The video clips are taken as input in each epoch with lr as 0.95. All the models are trained from scratch using FERV39k to present the benchmarks for 60 epochs with standard stochastic gradient descent (SGD) with momentum as 0.9 and uniformly sampled frame interval as 8.

Besides, as the number of sequences in FERV39k is limited for training, we exploit the data-augmentation techniques into the training set: randomly cropping, illumination changes and image flip. In order to reduce the dependence on the computation source, all cropped facial images are resized to $112\times112$, and whole images are resized to $112\times168$.


\noindent{\textbf{Evaluation metric.}} Following the standard practice \cite{W38_schuller2010cross, W7_mollahosseini2017affectnet, W12_jiang2020dfew} for evaluating FER or DFER, we choose two commonly used metrics: weighted average recall (WAR, also called as overall accuracy) and unweighted average recall (UAR). 

\subsection{Baseline Network}
According to the baseline architectures of action recognition in video \cite{W21_kay2017kinetics, W22_carreira2017quo,chen2021deep}, we first briefly define and describe several standard ConvNet architectures for DFER. We consider four typical approaches for DFER: 2D ConvNet, 2D ConvNet-LSTM on top of \cite{W39_yue2015beyond}, 3D ConvNet \cite{W26_tran2015learning, W40_ji20123d}, and Two-Stream 3D ConvNet. We then use these architectures as baselines and compare their performance by training and testing on whole FERV39k. Table \ref{tab:tab2} shows comparison results of four kinds of baseline architectures on FERV39k.

\noindent{\textbf{2D ConvNet.}}  The deep CNNs (2D ConvNet) such as VGG \cite{W41_simonyan2014very} and ResNet \cite{W42_he2016deep}, have made great success on image classification tasks \cite{W43_rawat2017deep}. Hence, we reuse them with minimal change for DFER. For processing a clip, features of all frames can be extracted and flatted into embeddings, which are concatenated and fed into classifier to obtain results.

\noindent{\textbf{2D ConvNet-LSTM.}} The structure of 2D ConvNet-LSTM is more appropriate for DFER by adding a recurrent layer to the model \cite{W44_donahue2015long} to introduce temporal information. Hence, we position an LSTM layer with 1024 hidden units and batch normalization layer (as proposed by Cooijmans et al. \cite{W4_dapogny2015dynamic}) after the last average pooling layer of 2D ConvNets. A fully connected layer is added on top as the classifier. 

\noindent{\textbf{3D ConvNet.}} 3D ConvNets(e.g. C3D \cite{W26_tran2015learning} and I3D \cite{W22_carreira2017quo}) can directly model hierarchical representations of spatio-temporal information with spatio-temporal (3D) filters. One issue with 3D ConvNets is that they have much more parameters than 2D ConvNets due to the additional kernel dimension. Besides, extra adjustment of network and output structures are required for DFER.

\begin{table*}[ht]
  \centering
\setlength\tabcolsep{0.008\linewidth}
\tiny
\begin{tabular}{lcccccccccccccc}
\toprule
Method         & All         & DL11k       & WIS9k       & SIA10k      & AI9k        & Social      & DailyLife   & Liveshow    & Talkshow    & Interview   & Contest     & Experiment  & Terror      & Crisis      \\ \midrule
R18            & 39.33/30.30 & 39.75/31.36 & 40.50/28.67 & 42.31/30.02 & 33.90/27.20 & 39.74/33.26 & 41.40/31.13 & 37.72/26.82 & 38.57/25.47 & 45.75/29.18 & 48.24/33.37 & 49.56/26.70 & 31.28/26.69 & 36.88/29.21 \\
R50            & 30.57/22.47 & 30.46/21.52 & 32.52/23.50 & 30.56/22.68 & 30.14/19.94 & 27.51/25.05 & 31.00/19.37 & 28.51/23.13 & 28.86/20.14 & 33.25/21.72 & 37.06/27.55 & 31.86/16.89 & 26.54/19.67 & 24.25/20.11 \\
VGG13          & 41.02/31.19 & 40.40/31.59 & 43.04/30.23 & 43.44/29.99 & 38.86/29.94 & 48.03/35.50 & 39.07/28.63 & 44.74/30.40 & 40.57/26.25 & 44.34/28.83 & 47.62/32.39 & 49.56/26.52 & 36.73/31.48 & 42.52/31.65 \\
VGG16          & 41.66/32.01 & 41.81/32.59 & 42.93/30.77 & 42.31/29.58 & 39.60/31.46 & 43.23/34.77 & 41.19/28.73 & 46.05/33.65 & 39.43/24.96 & 47.17/30.77 & 48.03/32.92 & 52.21/33.12 & 39.57/34.21 & 40.86/32.95 \\ \hline\hline
R18-LSTM       & 42.59/30.92 & 43.34/32.24 & 44.12/29.59 & 42.85/28.78 & 39.66/30.40 & 42.36/31.47 & 41.61/29.11 & 46.93/31.59 & 44.57/27.23 & 45.52/28.01 & 50.10/33.79 & 48.67/25.42 & 35.55/29.96 & 44.85/33.46 \\
R50-LSTM       & 40.75/32.12 & 40.93/32.91 & 41.74/30.70 & 42.16/30.39 & 38.01/31.16 & 42.79/35.70 & 41.61/28.00 & 40.35/30.40 & 40.00/27.42 & 43.87/30.02 & 48.24/34.32 & 47.79/33.17 & 36.26/32.46 & 39.87/31.87 \\
VGG13-LSTM     & 43.37/32.41 & 42.29/32.46 & 44.23/30.81 & 45.00/31.45 & 41.20/31.49 & 43.67/34.64 & 46.07/31.50 & 45.61/31.28 & 44.29/29.17 & 47.17/30.07 & 49.90/33.66 & 57.52/36.17 & 40.28/33.61 & 42.86/31.11 \\
VGG16-LSTM     & 41.70/30.93 & 42.99/32.32 & 41.63/28.42 & 43.83/29.83 & 37.04/29.39 & 49.34/36.83 & 44.37/30.58 & 36.84/25.76 & 41.14/26.39 & 46.23/27.39 & 48.65/34.15 & 53.10/30.03 & 36.26/32.83 & 41.53/33.59 \\ \hline\hline
C3D~\cite{W26_tran2015learning}            & 31.69/22.68 & 26.95/21.02 & 30.15/19.94 & 42.70/29.22 & 27.29/19.80 & 34.50/24.34 & 26.96/18.35 & 28.51/22.55 & 36.57/23.25 & 43.16/26.35 & 46.58/32.44 & 54.87/22.87 & 22.99/20.31 & 32.56/20.93 \\
I3D~\cite{W22_carreira2017quo}            & 38.78/30.17 & 38.56/29.25 & 38.52/29.11 & 40.55/31.07 & 37.44/28.15 & 37.55/32.05 & 39.70/26.09 & 37.72/30.57 & 26.29/18.27 & 41.51/27.87 & 45.55/35.43 & 53.10/31.56 & 33.89/29.10 & 36.54/28.81 \\
3D-R18~\cite{W45_tran2018closer}         & 37.57/26.67 & 37.69/27.47 & 38.40/24.85 & 40.40/26.08 & 33.45/25.40 & 41.48/29.83 & 35.67/24.95 & 39.04/25.12 & 36.29/21.86 & 42.69/22.70 & 44.10/28.32 & 54.87/32.50 & 31.28/27.83 & 37.21/27.25 \\ \hline\hline
Two C3D        & 41.77/30.72 & 41.45/31.37 & 43.44/29.77 & 44.71/30.15 & 37.89/28.09 & 47.16/32.22 & 35.46/23.26 & 41.23/25.74 & 42.00/27.89 & 46.23/28.45 & 48.03/32.31 & 63.72/37.55 & 35.78/30.47 & 40.86/29.60 \\
Two I3D        & 41.30/31.01 & 41.02/31.55 & 42.31/30.14 & 43.63/31.20 & 38.75/28.53 & 44.98/30.94 & 40.76/28.93 & 38.16/25.91 & 39.43/28.37 & 44.81/29.96 & 48.03/33.28 & 54.87/26.96 & 36.02/29.19 & 38.87/28.01 \\
Two 3D-R18     & 42.28/30.55 & 42.77/32.72 & 44.12/29.63 & 42.95/27.83 & 38.46/28.54 & 49.34/31.62 & 39.28/28.41 & 41.67/28.50 & 38.57/24.66 & 45.52/24.71 & 48.45/33.16 & 62.83/33.41 & 35.07/28.73 & 42.19/29.58 \\
Two R18-LSTM   & 43.20/31.28 & 42.20/31.66 & 44.91/30.37 & 46.33/31.09 & 40.40/30.04 & 47.60/35.60 & 40.55/27.09 & 44.74/26.55 & 43.43/27.52 & 47.41/28.50 & 53.00/33.93 & 57.52/24.56 & 36.49/29.94 & 43.85/31.45 \\
Two VGG13-LSTM & 44.54/32.79 & 44.65/32.96 & 45.25/31.45 & 46.57/31.88 & 40.63/30.96 & 48.03/36.43 & 46.92/31.55 & 48.25/33.02 & 45.14/28.30 & 46.70/28.35 & 52.80/35.32 & 53.98/31.66 & 37.44/32.49 & 46.84/35.11 \\ \hline\hline
Average        & 39.58/29.34 & 39.27/29.80 & 40.61/28.11 & 42.04/28.94 & 36.55/27.61 & 42.25/31.97 & 38.98/26.75 & 39.79/27.65 & 38.39/24.75 & 44.19/27.22 & 47.33/32.39 & 52.06/28.57 & 33.75/28.70 & 39.12/29.12 \\ \bottomrule
\end{tabular}
\vspace{-0.2cm}
\caption{Results of four kinds of baseline architectures trained from scratch on FERV39k (WAR/UAR).}
\label{tab:tab2}
\end{table*}

\begin{figure*}[ht]
  \centering
   \includegraphics[width=1.0\linewidth]{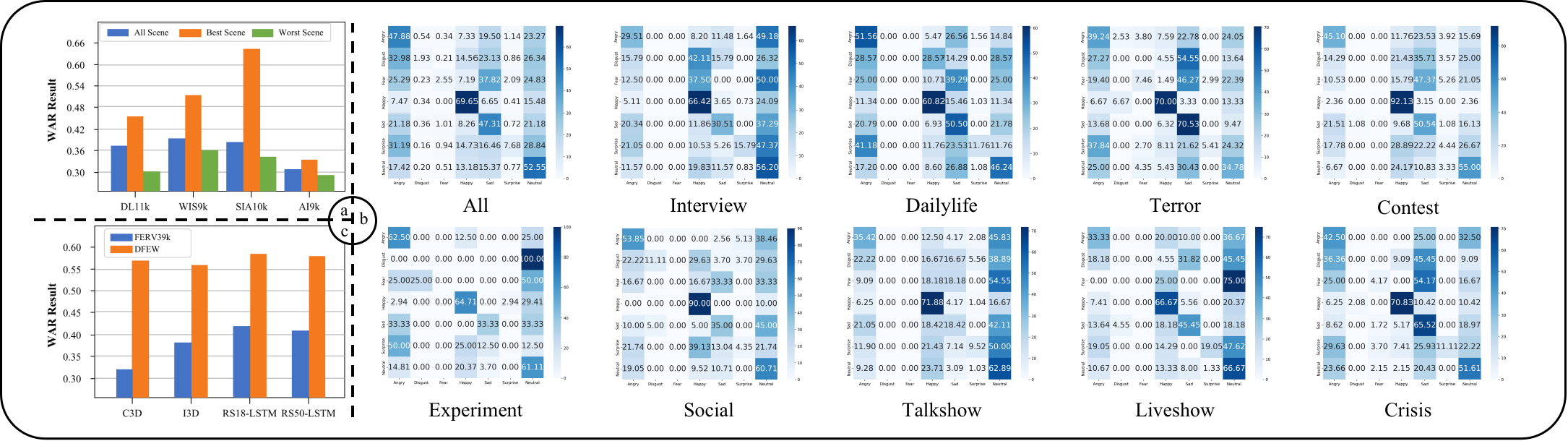}
   \vspace{-0.6cm}
   \caption{Further experiment analysis in detail. (a) The worst, average and best scene test result for RS50-LSTM trained on 4 scenarios respectively. (b) The confusion matrices of Two stream VGG13-LSTM (best performance) on the FERV39k and 9 representative scenes. (c) The comparison of performance on FERV39k and DFEW of 4 baseline methods.}
   \label{fig:fig7}
   \vspace{-0.5cm}
\end{figure*}

\noindent{\textbf{Two-Stream Networks.}}  Different from above methods, two-stream networks can encode the context components of scene, as well as facial expression of cropped facial image, together, inspired by CAER \cite{W19_lee2019context}. Specifically, we feed sequences of cropped face images and scene frames into the Two-Stream 3D ConvNets and 2D ConvNet-LSTM.

\subsection{Baseline Evaluation} 

On top of FERV39k, we systematically evaluate four kinds of baseline architectures across multiple scenes. Here we note that all training protocols follow the original papers unless stated otherwise. Table \ref{tab:tab2} shows results of four kinds baseline architectures on 9 representative scenes of FERV39k (showing WAR/UAR performance). All models are trained from scratch in experiments. 

In summary, performance on more specific scenarios (WIS9k, SIA10k) is better than others, and for 22 fine-grained scenes most of the methods achieve the highest result on Experiment (SIA10k) and lowest on Terror (AI9k). We attribute the results to the consistency \& intensity of expressions and the discriminability of spatial-temporal context features. Two stream 2D ConvNets-LSTM methods outperform the others where VGG13-LSTM has the best performance of 44.54\%. And 2D ConvNet-LSTM methods outperform the 3D ConvNet methods on both one and two stream structures. We recognize this as the LSTM has better global-local temporal feature utilization mechanism. In Section 4.4, we further explore the effect of scene \& method and challenges on FER39k for DFER.

\begin{figure*}[ht]
  \centering
   \includegraphics[width=1.0\linewidth]{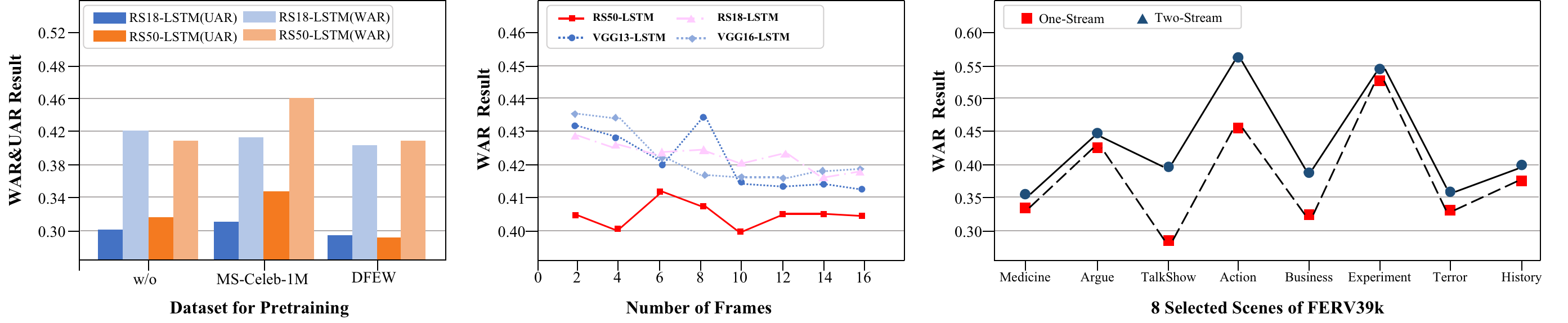}
   \vspace{-0.7cm}
   \caption{The result of three ablation studies. The charts from left to right are the result of pretraining effectiveness, sparse frame sampling effectiveness and scene information effectiveness, respectively.}
   \label{fig:fig8}
   \vspace{-0.5cm}
\end{figure*}

\noindent{\textbf{Cross-scenario challenge.}} We evaluate the cross-domain difficulty among 4 isolated scenarios via RS50-LSTM. Table \ref{tab:tab4} shows a nearly 8\% average cross-domain decline. And the largest decline of WIS9k experiment shows that it is more challenging to transfer model from weak interactive scenarios (e.g. WIS9k) to stronger ones than vice versa. To prove it, we also collect statistics of scene performance distribution of models training on corresponding scenario on Figure \ref{fig:fig7}(a). The result also shows WIS9k has both ideal performance and smaller difference among 4 scenarios. The result shows it a challenging task to overcome varieties of feature distribution of an expression in different domains of FERV39k. 

\begin{table}[ht]
\vspace{-0.1cm}
\setlength\tabcolsep{0.015\linewidth}
\small
\centering
\begin{tabular}{lcccc}
\toprule
\multirow{2}{*}{Source} & \multicolumn{4}{c}{Target}                            \\ \cline{2-5} 
                        & DL11k          & WIS9k         & SIA10k         & AI9k          \\ \midrule
DL11k                      & 37.69/27.21 & 29.98/19.93 & 31.15/21.87 & 24.27/18.54 \\
WIS9k                     & 27.04/19.95 & 40.5/26.6   & 31.78/19.9  & 24.62/19.24 \\
SIA10k                     & 28.57/21.92 & 31.39/19.95 & 39.72/24.9  & 27.75/20.28 \\
AI9k                      & 26.29/20.21 & 23.3/18.29  & 23.85/17.93 & 31.62/24.16 \\ \bottomrule
\end{tabular}
\vspace{-0.2cm}
\caption{Comparison of cross-scenario results on DL11k, WIS9k, SIA10k, and AI9k of FERV40k on RS50-LSTM.}
\label{tab:tab4}
\vspace{-0.35cm}
\end{table}

\noindent{\textbf{Scene difficulty and expression confusion.}} For further analyzing the difficulty in recognizing an expression in different scenes, we also provide the confusion matrices on Figure \ref{fig:fig7}(b) of selected scenes on the best performed network (VGG13-LSTM). The overall 10 matrices have similar distribution with sight offset among scenes in which method gains better performance on 4 obvious expressions and ‘Disgust’ is the hardest. The result shows an overall statics consistency with previous datasets e.g. DFEW. However, some subtle changes are worth to be noticed. For example, the performance for 'Sad' declined in Talkshow, Liveshow and Experiment as well as 'Angry' in Interview and Contest. This situation may caused by the changes of expressions in intensity, feature and frequency of occurrence (long-tailed distribution) in specific scenes. There are obvious bias and heterogeneity in our built FERV39k, which make it a challenging dataset. We summarize several directions might work: (1) Long temporal modeling; (2) Scene reasoning; (3) Global-local fusion in spatial and temporal.

\noindent{\textbf{Comparing performance with the existing dataset.}} To emphasize the difficulty of FERV39k, we compare results with three baseline architectures on DFEW (without two stream baseline). The methods on DFEW get higher average results of about 10\% (Figure \ref{fig:fig7}(c)), which proves FERV39k is more challenging to state-of-the-art methods. We attribute this to the following reasons: a) FERV39k tripled the number of clips than DFEW, b)Data variety represents a real-world challenge for existing algorithms and c) 22 scenes in this dataset require further application-oriented researches. 


\subsection{Ablation Studies}


\noindent{\textbf{Does pre-training on large-scale datasets help?}} We employ RS18-LSTM and RS50-LSTM with and without pretraining using MS-Celeb-1M \cite{guo2016ms} and DFEW \cite{W12_jiang2020dfew} our built FERV39k. The experiment shows that the former ones are not obviously outperform the latter in Figure \ref{fig:fig8} (left). One potential reason is that the scene and feature distribution of FERV39k is different from other datasets.

\noindent{\textbf{Is sparse sampling sufficient for DFER?}} The sparse sampling schemes\cite{W34_shao2020finegym, W49_caba2015activitynet, gowda2021smart} often lead to high efficiency and promising accuracy in action recognition. In order to explore whether sparse sampling is sufficient for DFER, we further investigate the influence of sampling frames on DFER performance. Here we adjust the number of input frames from 2 to 16 in steps of 2 for four 2D ConvNet-LSTM networks on FERV39k. The results in Figure \ref{fig:fig8} (middle) show that the performance trend varies among different methods but as frames increase over a threshold, the effect tends to be flat or fluctuating decline slightly. These results also show that more subtle sampling method should be used and the key frames extraction might be a point \cite{ghodrati2021frameexit, zheng2020dynamic}.

\noindent{\textbf{Is scene information auxiliary for DFER?}} In order to further understand whether the scene information can boost the performance of DFER methods, we compare Two and single stream I3D network on FERV39k benchmark. We select the best and worst result scene of 4 scenarios and provide results in Figure \ref{fig:fig8} (right), which show that two-stream networks can enhance the face-only model and achieve better result in most scenes due to the fusion of the context information in scene. For example, we could easily guess the expression as “Sad” with the facial region and scene contexts when someone comes.

\noindent{\textbf{Why current methods fail to handle FERV39k?}}  By carefully summarizing all the experiments, we conclude some factors that make FERV39k challenging to four baseline architectures: (1) Limited expression-related frames, especially scenes with frequent emotional changes. (2) Subtle spatial semantics, which involve differences in face and scene-face relationships. (3) Complex temporal dynamics, such as the direction of motion, and the degree of rotation. In addition, the FERV39k dataset poses higher requirements for intermediate representation which is hard to be extracted due to the diversity in one scene.






\section{Conclusion}
\label{sec:Conclusion}

In this paper, we build a large-scale multi-scene dataset (FERV39k) for FER in videos. Compared with existing video-based datasets, our FERV39k has many distinctive characteristics: 1) Automatic generation of large-scale candidate video clips;  2) Well-designed workflow of crowdsourcing and professional annotation for high-quality data labeling; 3) Raising four kinds of challenges and difficulties for FER in videos; 4) Application-oriented multi-scene hierarchy for the robustness of DFER methods. To benchmark the FERV39k, we design four kinds of baseline architectures for video-based FER and give an in-depth evaluation and ablation studies. These results present some important challenges and uncover critical messages for advancing the area of video-based FER in the future.

\section{Acknowledgments}
This work was supported by National Natural Science Foundation of China (No.62072112), National Key R\&D Program of China (2020AAA0108301),  Scientific and Technological Innovation Action Plan of  Shanghai Science and Technology Committee (No.20511103102), Fudan University-CIOMP Joint Fund (No. FC2019-005), Double First-class Construction Fund (No. XM03211178), and partly supported by the National Natural Science Foundation of China under Grant Nos. 62106051 and the Shanghai Pujiang Program Nos. 21PJ1400600.

{\small
\bibliographystyle{ieee_fullname}
\bibliography{main}
}

\newpage
\clearpage
\begin{center}
	\textbf{\Large Appendix}
\end{center}

\appendix

\section{Annotation Documentation}
\label{sec:Annotation}

In this paper, we build a large-scale multi-scene dataset (FERV39k) for FER in videos. There are four isolated scenarios subdivided into 22 scenes: 6 scenes \{Argue, Social, School, Medicine, Conflict, and Daily-Life\} designed for Daily Life (DL11k), 6 scenes \{Action, ScholarReports, Speech, Elegant-Art, Live-Show, and Talk-Show\} for Weak-Interactive Shows (WIS9k), 6 scenes \{Business, Experiment, Official-Event, Crime, Interview, Contest\} for Strong-Interactive Activities (SIA10k) and 4 scenes \{History, Terror, War and Crisis\} for Anomaly Issues (AI9k). It tends to be ambiguous and remains as a big challenge to build DFER datasets with the complexity and variability of spatial-temporal dynamics. To ensure the consistency of annotation of 7 basic expressions in different scenes, we write a handbook (also called annotation documentation) to clarify the definition of 7 basic expressions (“Angry”, “Disgust”, “Fear”, “Happy”, “Sad”, “Surprise”, and “Neutral” ). Afterwards, we formulate representative examples with 7 basic expressions across 22 scenes.

\subsection{Daily Life (DL11k)}

DL11k scenario is composed of 6 scenes commonly experienced in the daily life. In this scenario, scenes vary from each other a lot due to the complexity of real-life activities. Figure \ref{fig:SDL11k} shows an overview of 7 expressions in 6 scenes.

\subsection{Weak-Interactive Shows (WIS9k)}

WIS9k scenario is composed of 6 kinds of shows. The person in scenes of WIS9k usually maintains a consistent emotional state over a long period of time. Besides, the intensity of expressions is much higher. Figure \ref{fig:SWIS9k} shows an overview of 7 expressions in 6 show scenes.

\subsection{ Strong-Interactive Activities (SIA10k)}

SIA10k scenario mainly focuses on activities with strong interaction. In these scenes, the emotion of a person is usually influenced by other people and environment. As a result, the distribution of expressions shows great instability and diversity. Figure \ref{fig:SSIA10k} shows an overview of 7 expressions with great diversity in 6 scenes.

\subsection{Anomaly Issues (AI9k)}

AI9k scenario contains 4 hardly seen scenes in our daily life. It is difficult for both researchers and DFER methods to distinguish unexpected appearances and changes of an expression in these scenes. Figure \ref{fig:SAI9k} shows some unusual appearances with 7 basic expressions.

\section{Generation of Candidate Video Clips}
\label{sec:Related_Work}

After reviewing top-level 22 scenes, we collect scenes corresponding online videos, TV shows and movies from open search engines. The first step of building a dynamic dataset is candidate video clip selection. The main problem for us is to acquire available full context and single face clips. Existing works ask annotators to manually segment video clips with expressions via video editing software. However, it is costly to do so when segmenting many raw videos into several qualified video clips among 0.5$\sim$4 seconds. Hence, we design a four-stage strategy to collect and generate candidate video clips.

\subsection{Rule-based Selection Mechanism}
\begin{figure}[ht]
  \centering
   \includegraphics[width=1.0\linewidth]{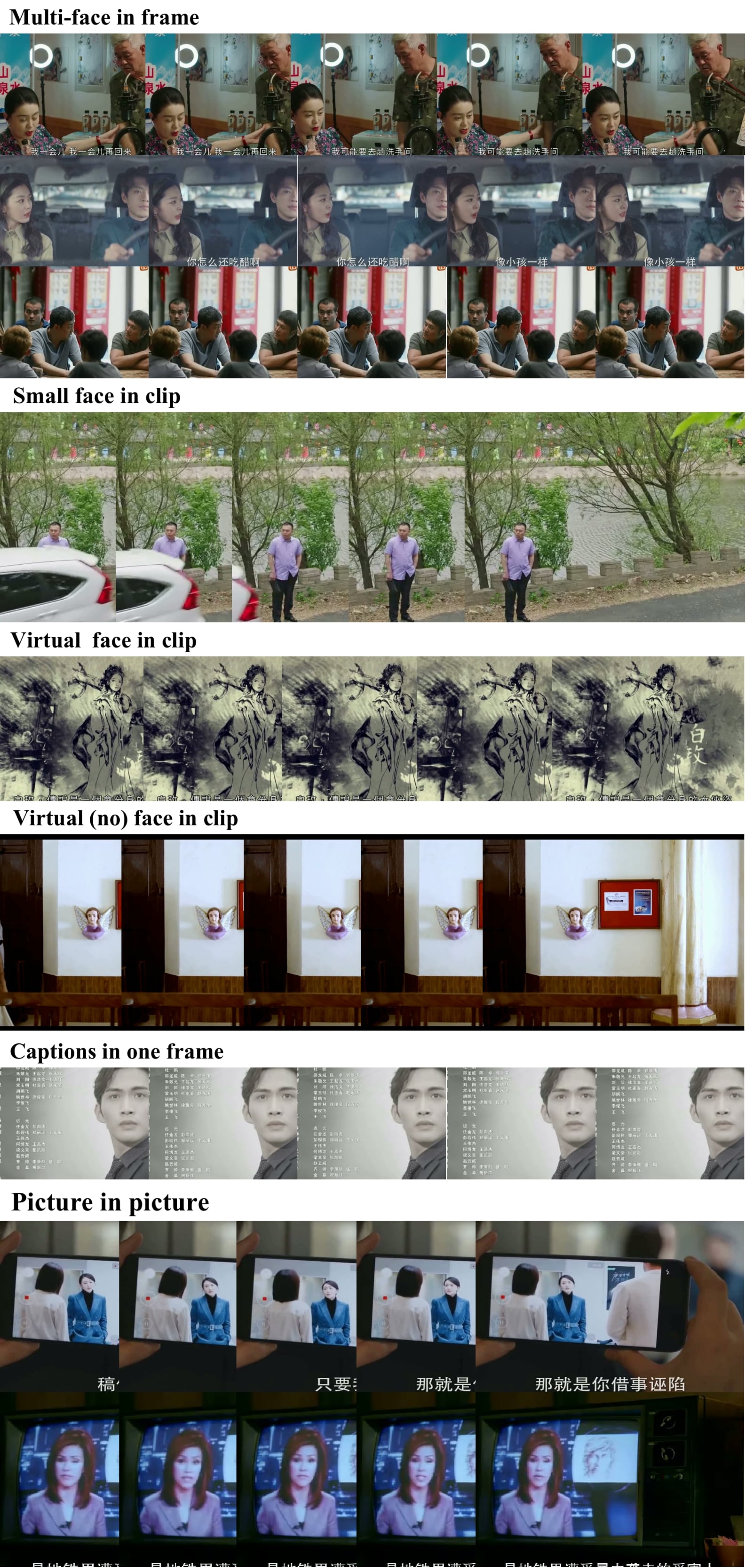}
   \vspace{-0.6cm}
   \caption{Six kinds of video clips need to be removed due to violating the rule of only one face in one frame.}
   \label{fig:rules1}
    \vspace{-0.5cm}
\end{figure}

\begin{figure}[ht]
  \centering
   \includegraphics[width=1.0\linewidth]{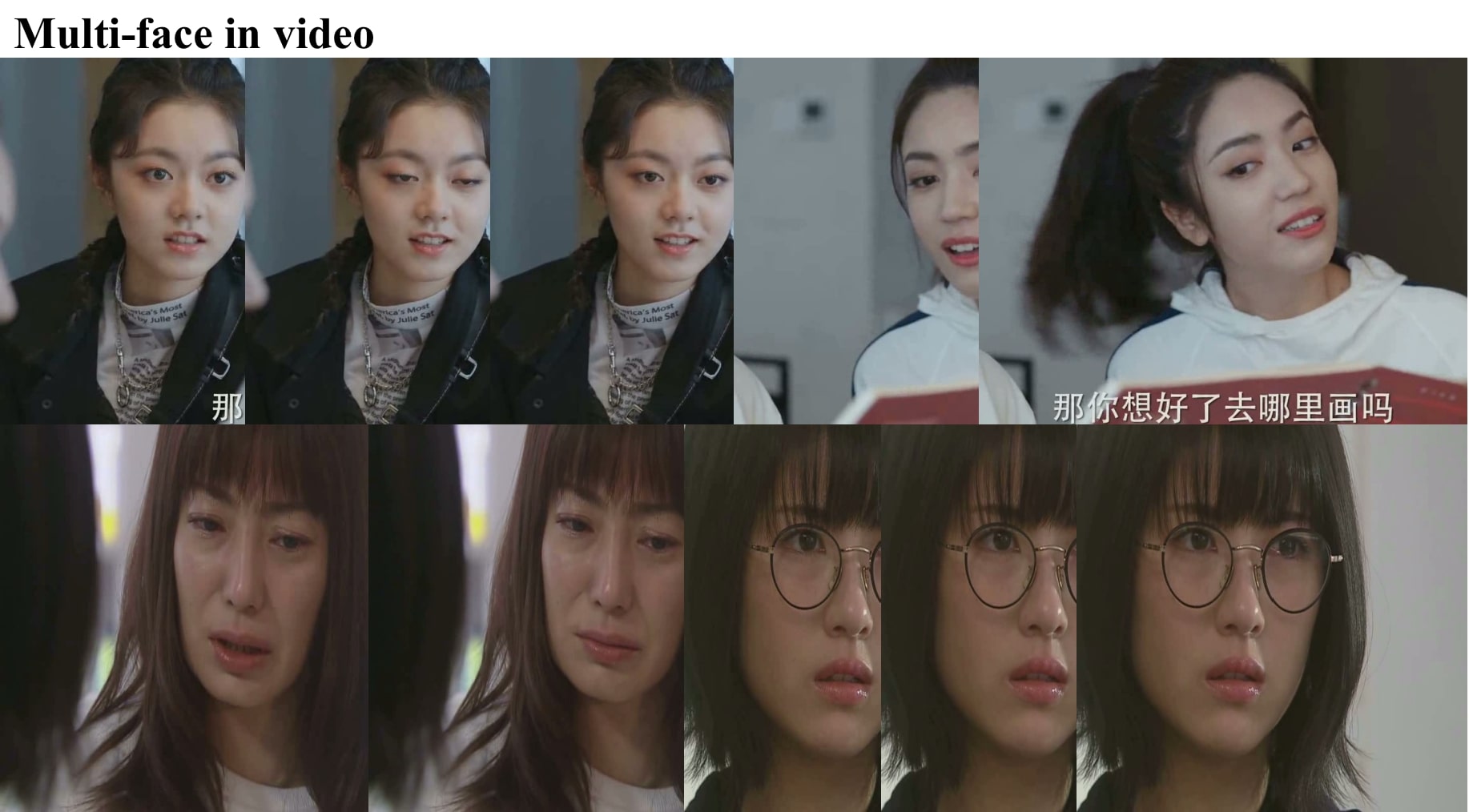}
   \vspace{-0.6cm}
   \caption{Multi-face in one video.}
   \label{fig:rules2}
    \vspace{-0.5cm}
\end{figure}

\begin{figure*}[ht]
  \centering
  \includegraphics[width=1.0\linewidth]{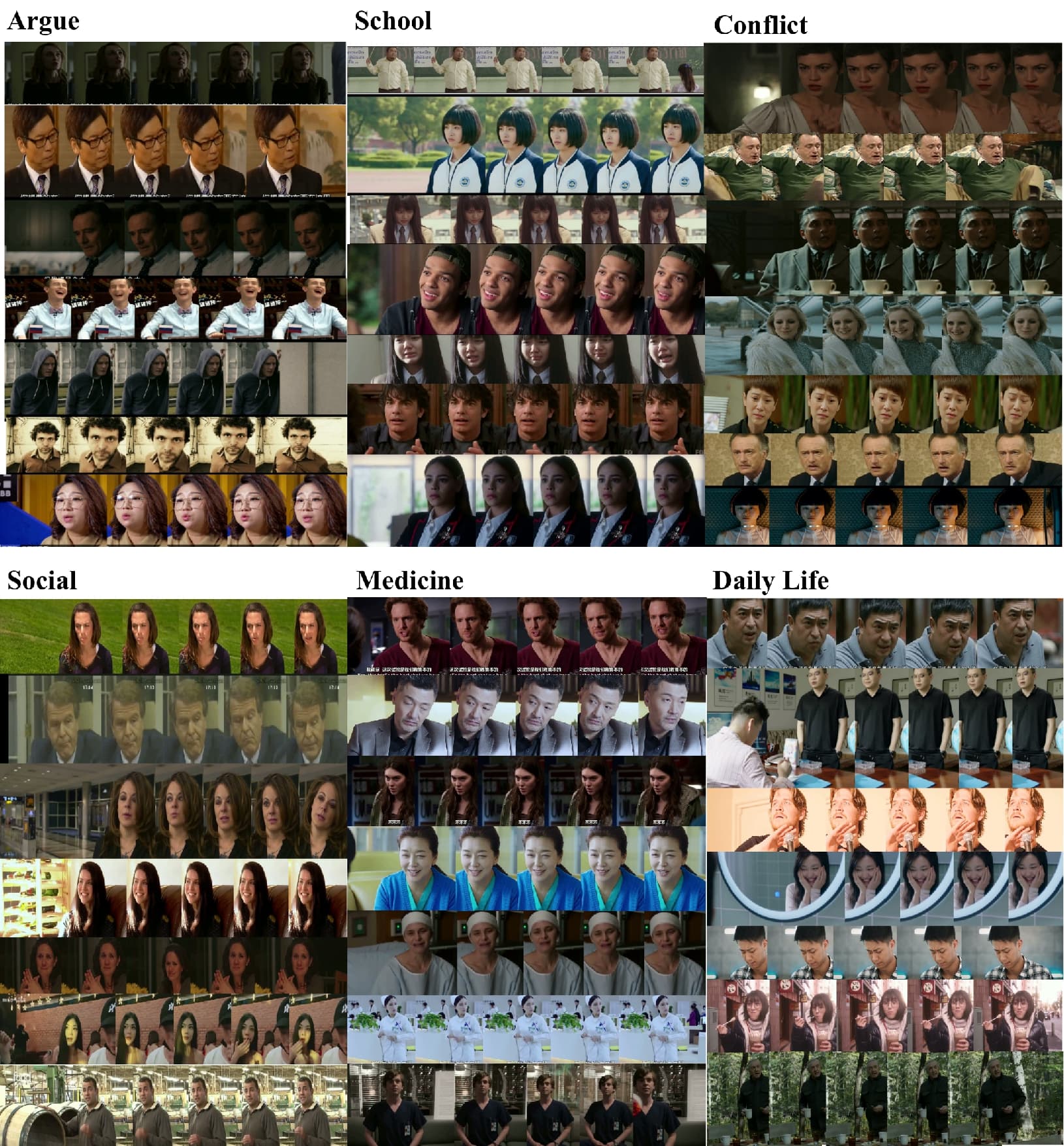}
  \vspace{-0.6cm}
  \caption{7 expressions in 6 scenes of DL11k. Each scene show 7 representative frame-level video clips with Angry, Disgust, Fear, Happy, Sad, Surprise, and Neutral expression from top to bottom.}
  \label{fig:SDL11k}
\end{figure*}

\begin{figure*}[ht]
  \centering
  \includegraphics[width=1.0\linewidth]{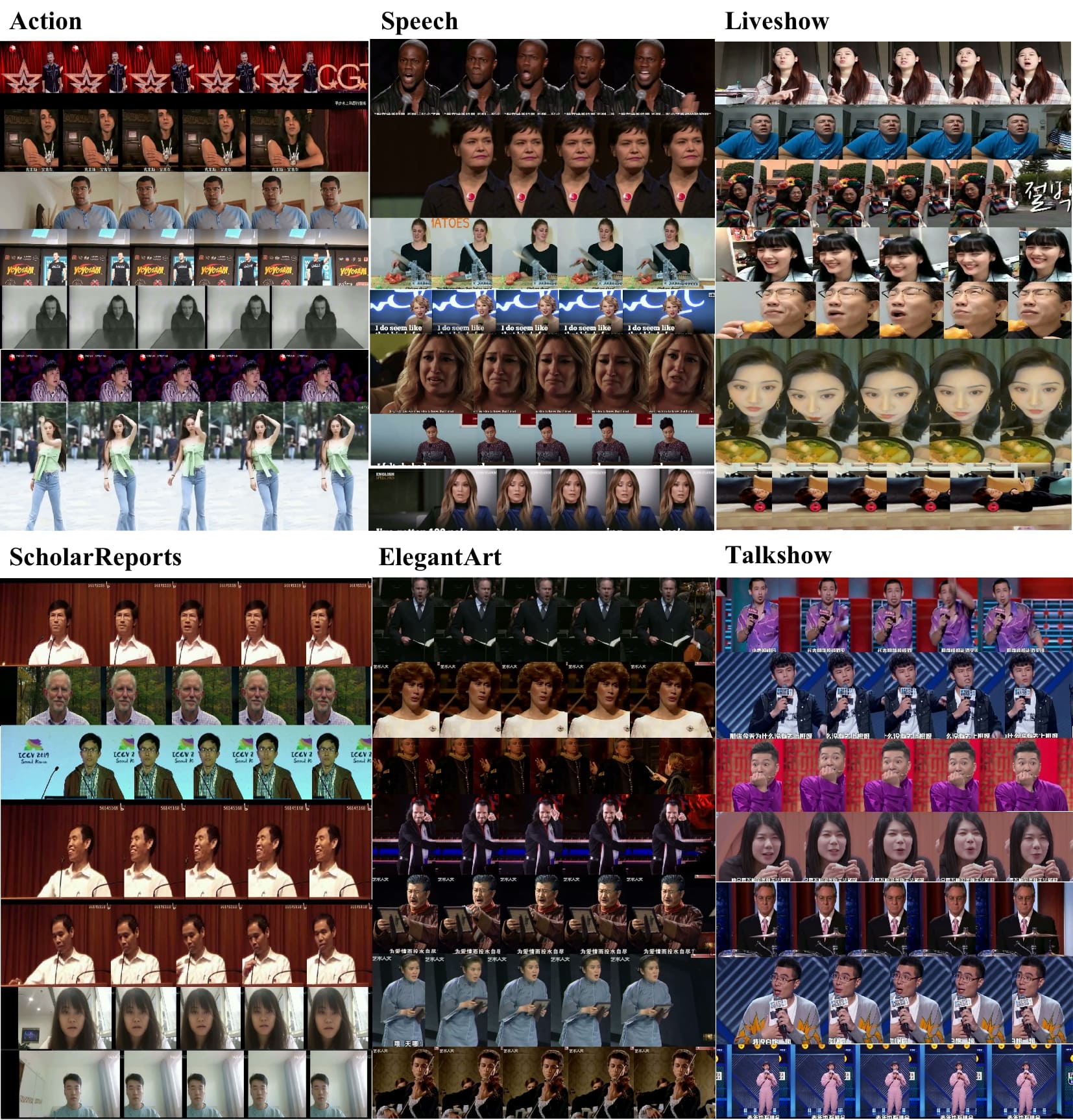}
  \vspace{-0.6cm}
  \caption{7 expressions in 6 scenes of WIS9k. Each scene show 7 representative frame-level video clips with Angry, Disgust, Fear, Happy, Sad, Surprise, and Neutral expression from top to bottom.}
  \label{fig:SWIS9k}
\end{figure*}

\begin{figure*}[ht]
  \centering
  \includegraphics[width=1.0\linewidth]{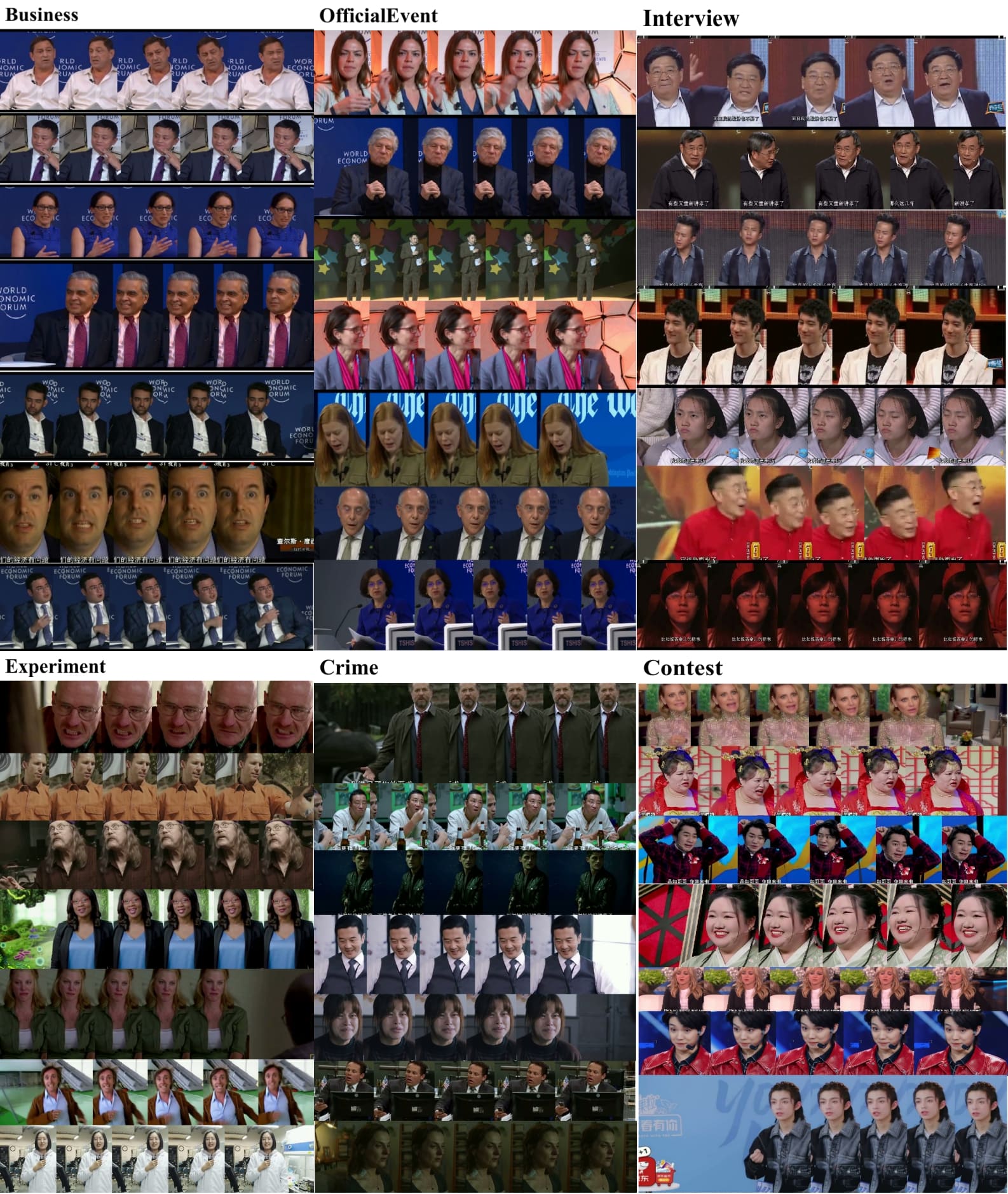}
  \vspace{-0.6cm}
  \caption{7 expressions in 6 scenes of SIA10k. Each scene show 7 representative frame-level video clips with Angry, Disgust, Fear, Happy, Sad, Surprise, and Neutral expression from top to bottom.}
  \label{fig:SSIA10k}
\end{figure*}

\begin{figure*}[ht]
\vspace{-1cm}
  \centering
  \includegraphics[width=1.0\linewidth]{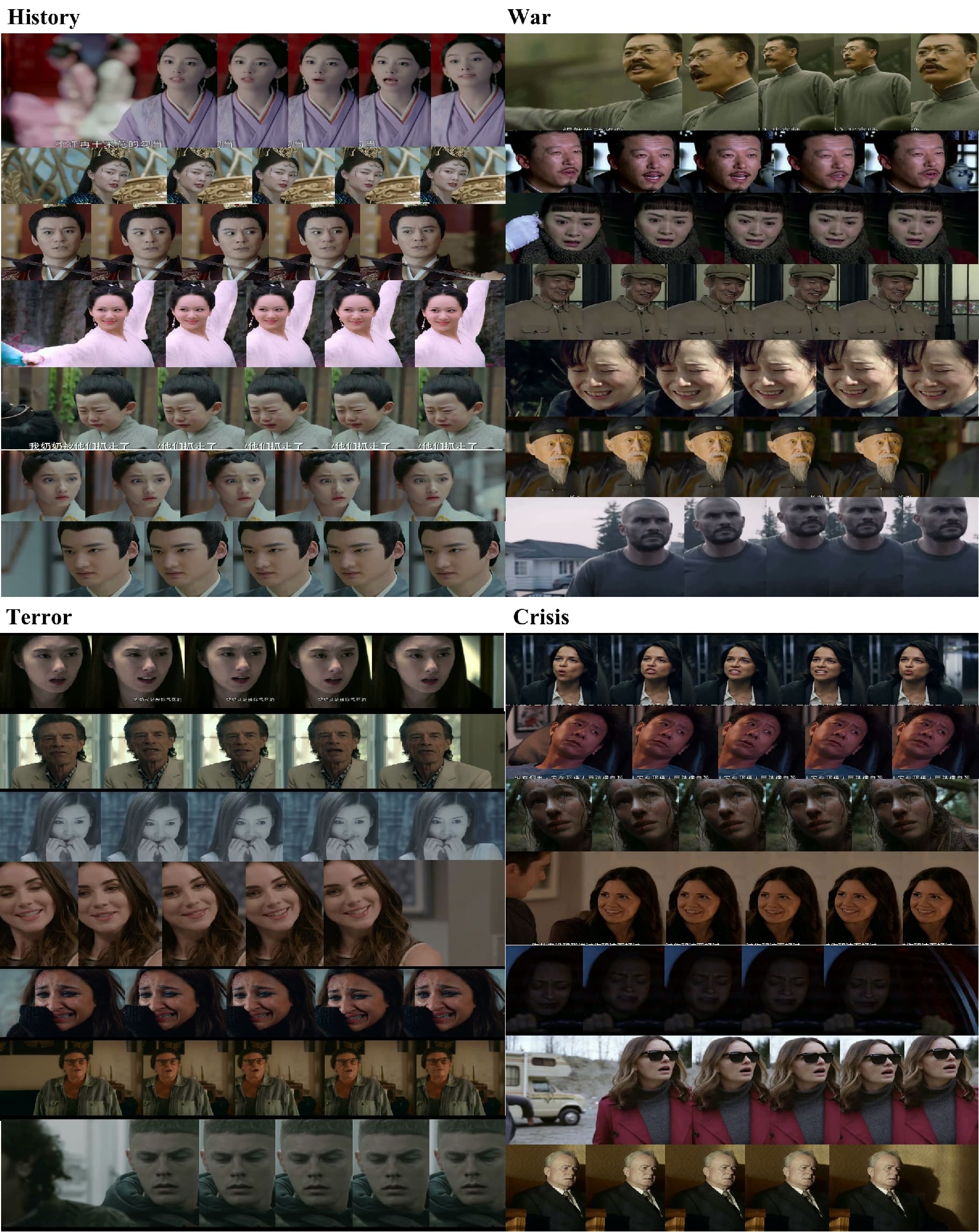}
  \vspace{-0.6cm}
  \caption{7 expressions in 4 scenes of AI9k. Each scene show 7 representative frame-level video clips with Angry, Disgust, Fear, Happy, Sad, Surprise, and Neutral expression from top to bottom.}
  \label{fig:SAI9k}
\end{figure*}

At first, we randomly split each raw video into many video clips among 0.5$\sim$4 seconds according to this work \cite{W36_ben2021video}. It is impossible to annotate millions of clips randomly generated from thousands of raw videos. Hence, we need to make some rules to discard even $\frac{19}{20}$ raw clips in some scenes such as Interview, Speech, etc. In addition, the selected clips tend to contain many samples with same person. In order to further generate finer candidate video clips, we make a rule list to help our well-designed mechanism to adaptively select satisfactory clips from a twenty-fold amount of clips than expected scale of final dataset. The rule list is included as:

\begin{itemize}
\setlength{\itemsep}{0pt}
\setlength{\parsep}{0pt}
\setlength{\parskip}{0pt}

\item Our algorithm can automatically segment a raw video into many video clips containing only one face lasting for 0.5$\sim$4 seconds by removing such clips: multi-face, small face, vitural face, captions and picture in picture. Some examples in Figure \ref{fig:rules1} show that six kinds of such video clips need to be removed by both our algorithm and annotators.
\item Only one person can appear in one video clip, no other people are allowed to appear. We use a face matching technique to completely remove such video clips with identity variation in a frame sequence. Some examples in Figure \ref{fig:rules2} show that such video clips are expcted to be removed by both our algorithm and annotators.
\item More than 90\% frames in one video clip must contain detectable faces of the same person.
\item The total clip latency should follow the previous statistic researches and the number of selected clips from each video should follow an average total duration distribution. A further random selection method is used to fulfill this purpose.
\end{itemize}

\subsection{FER-based detector}
According to the above rule-based algorithm, we can find that the majority expressions of video clips generated from the raw videos are Neutral and Happy. In order to balance the data distribution, we train a well-designed and light-weight FER detector with high accuracy for recognizing in-the-wild facial expressions. As Real-world Affective Face Database (RAF-DB) \cite{li2017reliable} contains about 30,000 real-world facial images, we use RAF-DB to train our FER detector based on ResNet50 model. The FER-based detector is implemented in PyTorch-GPU using GeForce RTX 2080ti GPUs. The learning rate $lr$ is initialized to 1e-3, and $lr$ is exponentially decayed by a factor of 0.95 every epoch. The stochastic gradient descent (SGD) is optimized with 0.9 Momentum and 1e-3 weight decay. Batch size is fixed at 32. The maximum epoch number is 120. Finally, the FER-based detector achieves the overall accuracy of 87.53\% of 7 basic expressions, which shows the comparable performance with other state-of-the-art methods \cite{W1_wang2020region,li2021adaptively} in limited computation sources. As a result, we utilize this FER-based detector to refine these clips with threshold value to generate relatively balanced candidate video clips for 7 basic expressions.


\section{Annotation Workflow }
\label{sec:Agreement}

In our designed procedure, there are two roles named crowd-sourcing annotators (20 workers) and professional researchers (10 workers), respectively. The crowd-sourcing annotators are relatively cost-less but with lower reliability. And the professional researchers cost more for labeling a clip but with the highest reliability (over the crowd-sourcing annotators, statistics-based inspection methods and even the administrators). Before annotating candidate video clips, two teams are both provided with a handbook. And the personalized designed platform is used for annotation. Five steps and two stages (annotation and judgement) in total are designed for the annotation work-flow:

\begin{itemize}
\setlength{\itemsep}{0pt}
\setlength{\parsep}{0pt}
\setlength{\parskip}{0pt}

\item Data grouping and flag presetting (annotation stage): Before annotation, clips are randomly divided into several groups and 5\% of clips in each group are annotated by the professional researchers and mixed into the raw materials (labels are hidden to crowd-sourcing annotators). Afterwards, the mixed groups are copied for 3 times. The copies of a specific group have unique group IDs which are invisible to annotators. Then we randomly shuffle the grouped materials and provide them to the crowd-sourcing annotators.
\item Annotation (annotation stage): The annotators are asked to choose the most likely expression from 26 fine-grained labels on the platform. Each label corresponds to an expression in “Angry”, “Disgust”, “Fear”, “Happy”, “Sad”, “Surprise” and “Neutral” and the chosen label can be automatically converted through the platform. For the samples without proper correspondence, annotators can press 'PASS' and the clips will be marked as illegal.
\item Auto-checking via error statistics (judgement stage): With the pipeline of annotation, the annotated materials are firstly checked via the flag-recapture based error statistics method \cite{bell1974population}. In this step, we collect all the groups and calculate correct rate of preset check data as the total correct rate.  We design a two-level threshold of 40\% and 80\%. The group with correct rate lower than 40\% will be marked as unacceptable and retreated to crowd-sourcing annotators; The group with correct rate between 40\% and 80\% will be marked as improper with additional warnings and passed to professional judgment; And the group with over 80\% correct rate will be marked as accept and passed to professional judgment without notes.
\item Professional judgment (judgement stage): Instead of choosing the corresponding expression from the multiple labels, the professional researchers only need to decide whether the labels are proper to the clips and whether the annotation of a group is reliable. The unreliable (unaccepted) ones will also be retreated but with extra instructions, and the improper ones will be relabeled by the professional researchers as the supplementary annotation, and a notice and advice will feedback to crowd-sourcing annotators.
\item Final decision (judgement stage): A weighted winner-takes-all (WWTA) voting mechanism is used. During the voting, supplementary annotation have twice higher weight than the normal annotation which means that, the professional researchers have higher confidence but can be refuted by a broader consensus. 
\end{itemize}

\section{Agreement}
\label{sec:Agreement}
\begin{itemize}
\setlength{\itemsep}{0pt}
\setlength{\parsep}{0pt}
\setlength{\parskip}{0pt}

\item The FERV39k dataset is available to \textbf{non-commercial research purposes} only.
\item All videos of the FERV39k dataset are obtained from the Internet which are not property of ***. Our group is not responsible for the content nor the meaning of these videos.
\item You agree \textbf{not to} reproduce, duplicate, copy, sell, trade, resell or exploit for any commercial purposes, any portion of the videos and any portion of derived data including but not limited to frames, and cropped face images 
\item You agree \textbf{not to} further copy, publish or distribute any portion of the FERV39k dataset. Except, for internal use at a single site within the same organization it is allowed to make copies of the dataset.
\item Our group reserves the right to terminate your access to the FERV39k dataset at any time.
\end{itemize}

\section{More Results of Comparisons and Confusion Matrices Under Different Scenes}
\label{sec:Results}
 On top of the FERV39k, we systematically evaluate four kinds of baseline architectures following action recognition baselines \cite{W21_kay2017kinetics, liu2021no, W22_carreira2017quo} and investigate inter-scene and intra-scene performances based on RS50-LSTM network, we design two kinds of experimental schemes: 1) Training all baselines on all data collected from 22 scenes, and test on the whole dataset, four scenarios, and each scene. 2) Training the RS50-LSTM networks on four scenarios, respectively, and test on four scenarios, and their sub-scenes. We present more results of confusion matrices of different methods, i.e. RS18-LSTM, C3D, Two RS18-LSTM, and Two VGG13-LSTM (all networks training from scratch) on four scenarios, i.e., DL11k, WIS9k, SIA10k, and AI9k in Figure \ref{fig:DL11k}, Figure \ref{fig:WIS9k}, Figure \ref{fig:SIA10k}, and Figure \ref{fig:AI9k}, respectively. According to the results, We observe that existing four kinds of baseline architectures fail to perform well in distinguishing challenging expressions such as 'Fear' and 'Disgust' with variable intensities across different scenes.

\subsection{Daily Life (DL11k)}
\begin{figure*}[ht]
  \centering
   \includegraphics[width=1.0\linewidth]{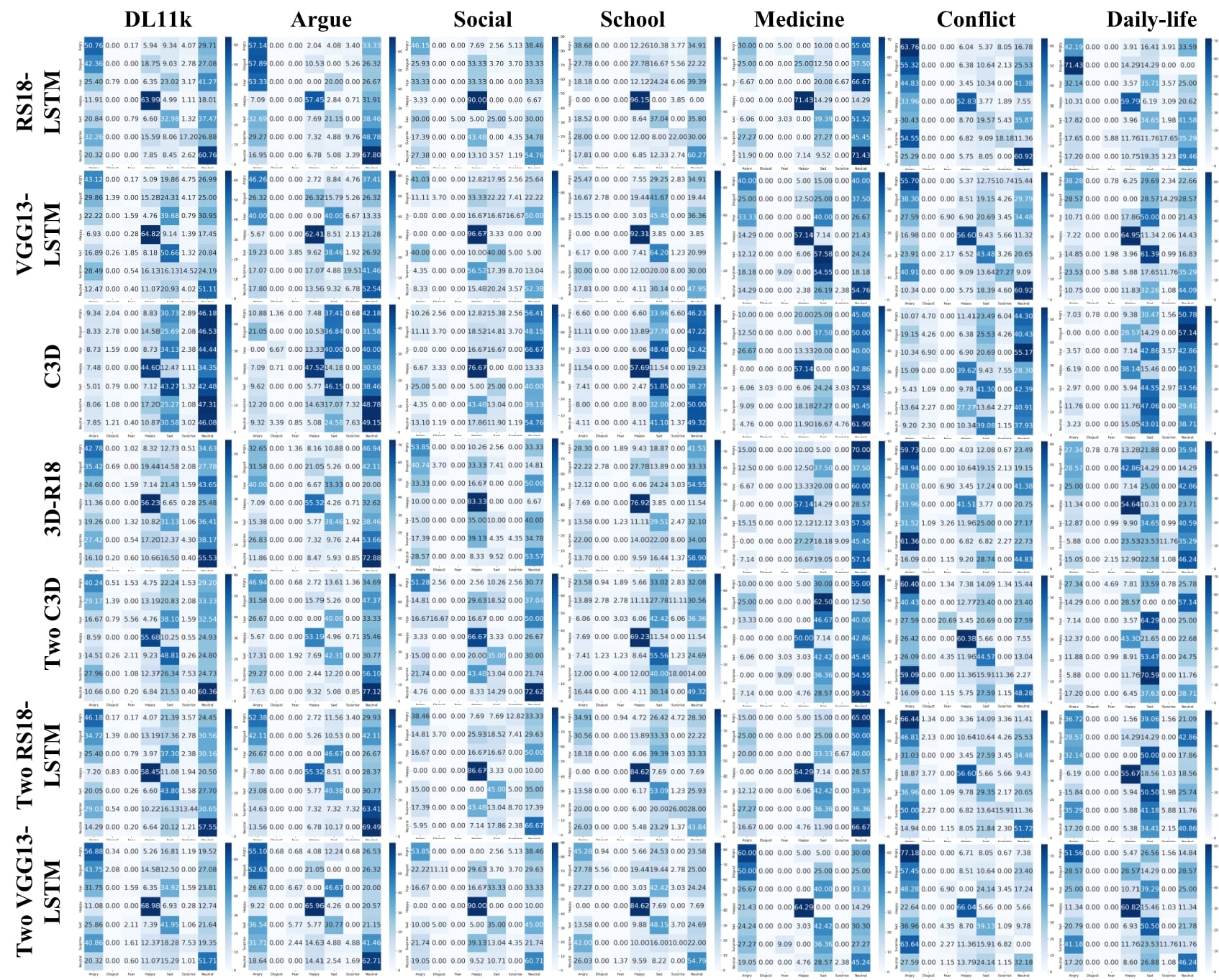}
   \vspace{-0.6cm}
   \caption{Confusion matrices of different methods on all scenes and then testing on DL11k, and its sub-scenes.}
   \label{fig:DL11k}
\end{figure*}

Table \ref{tab:dl_11k} shows comparison of four kinds of baseline architectures training on all scenes and then testing on all scenes, DL11k, and its sub-scenes. The confusion matrices of this experiment is also provided in Figure \ref{fig:DL11k}. And Table \ref{tab:dl_11k_4_to_27} shows the experiment result of intra-scenario performance consistency and inter-scenario invariance via RS50-LSTM training on four scenarios and testing on DL11k, and its sub-scenes. 

\subsection{Weak-Interactive Shows (WIS9k)}
\begin{figure*}[ht]
  \centering
   \includegraphics[width=1.0\linewidth]{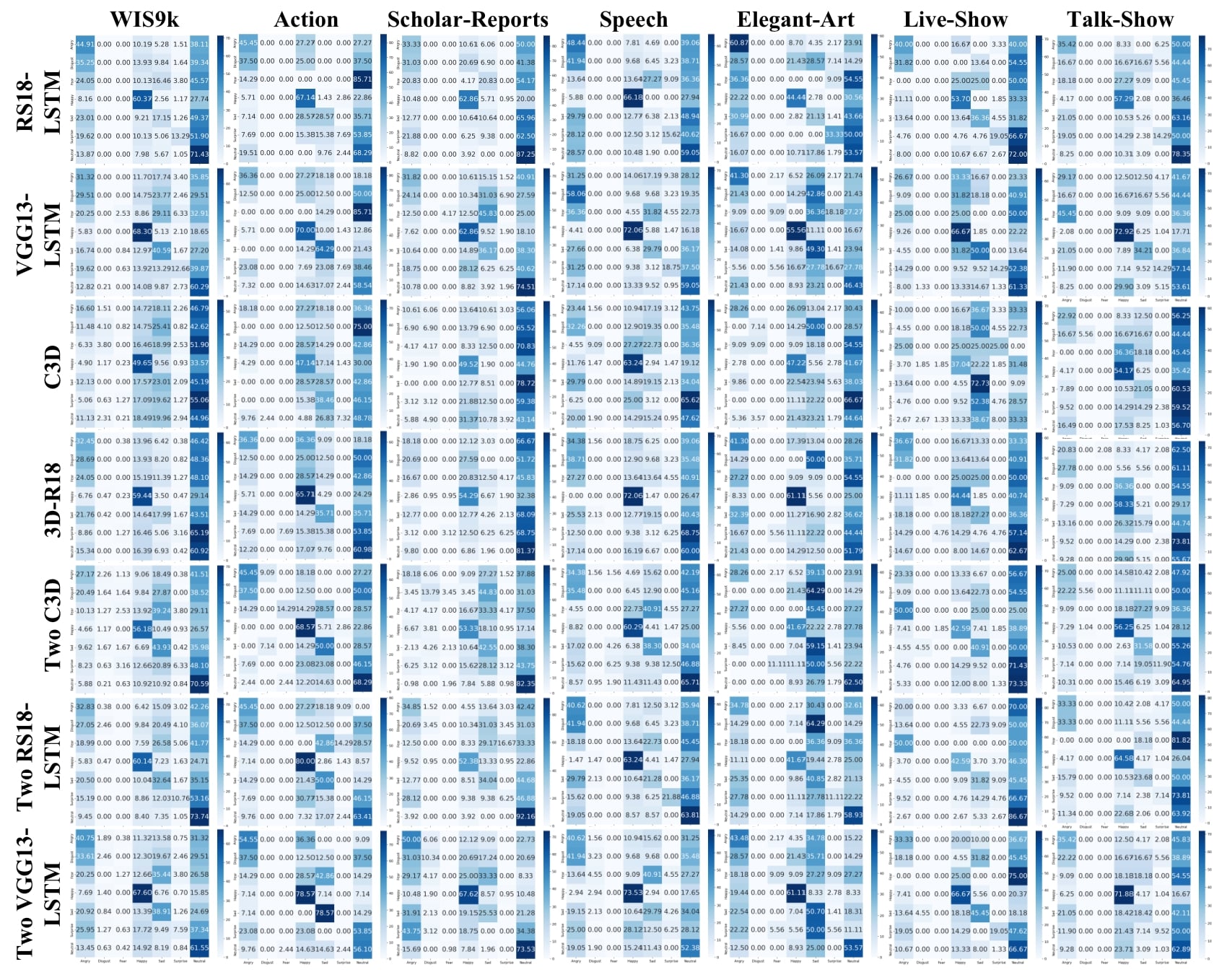}
   \vspace{-0.6cm}
   \caption{Confusion matrices of different methods on all scenes and then testing on WIS9k, and its sub-scenes.}
   \label{fig:WIS9k}
\end{figure*}

Table \ref{tab:wis_9k} shows comparison of four kinds of baseline architectures training on all scenes and then testing on all scenes, WIS9k, and its sub-scenes. The confusion matrices of this experiment is also provided in Figure \ref{fig:WIS9k}. And Table \ref{tab:wis_10k_4_to_27} shows the experiment result of intra-scenario performance consistency and inter-scenario invariance via RS50-LSTM training on four scenarios and then testing on WIS9k, and its sub-scenes. 

\subsection{ Strong-Interactive Activities (SIA10k)}
\begin{figure*}[ht]
  \centering
   \includegraphics[width=1.0\linewidth]{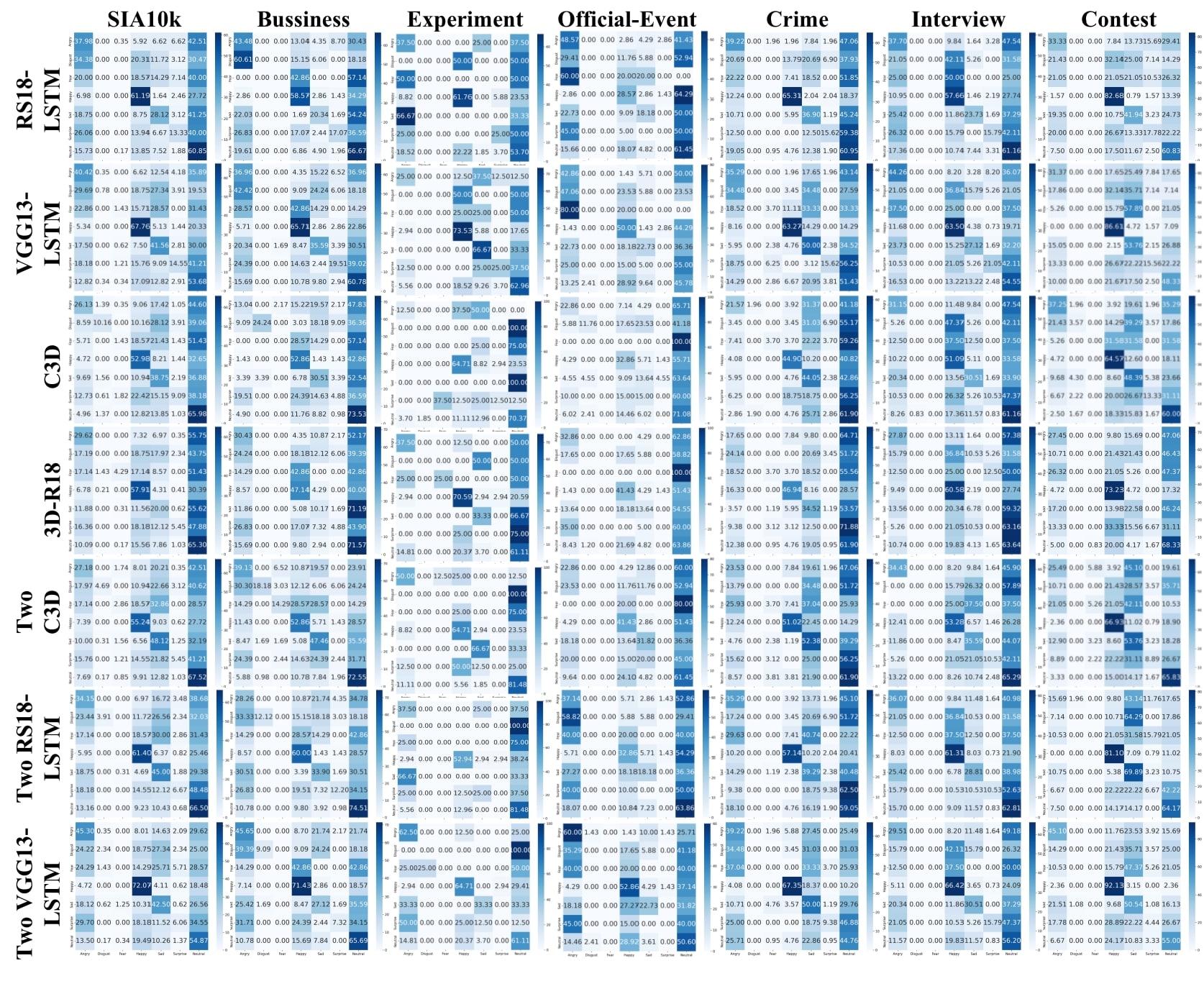}
   \vspace{-0.6cm}
   \caption{Confusion matrices of different methods training on all scenes and then testing on SIA10k, and its sub-scenes.}
   \label{fig:SIA10k}
\end{figure*}

Table \ref{tab:sia_10k} shows comparison of four kinds of baseline architectures training on all scenes and then testing on all scenes, SIA10k, and its sub-scenes. The confusion matrices of this experiment is also provided in Figure \ref{fig:SIA10k}. And Table \ref{tab:sia_10k_4_to_27} shows the experiment result of intra-scenario performance consistency and inter-scenario invariance via RS50-LSTM training on four scenarios and then testing on SIA10k, and its sub-scenes. 

\subsection{Anomaly Issues (AI9k)}
\begin{figure*}[ht]
  \centering
   \includegraphics[width=1.0\linewidth]{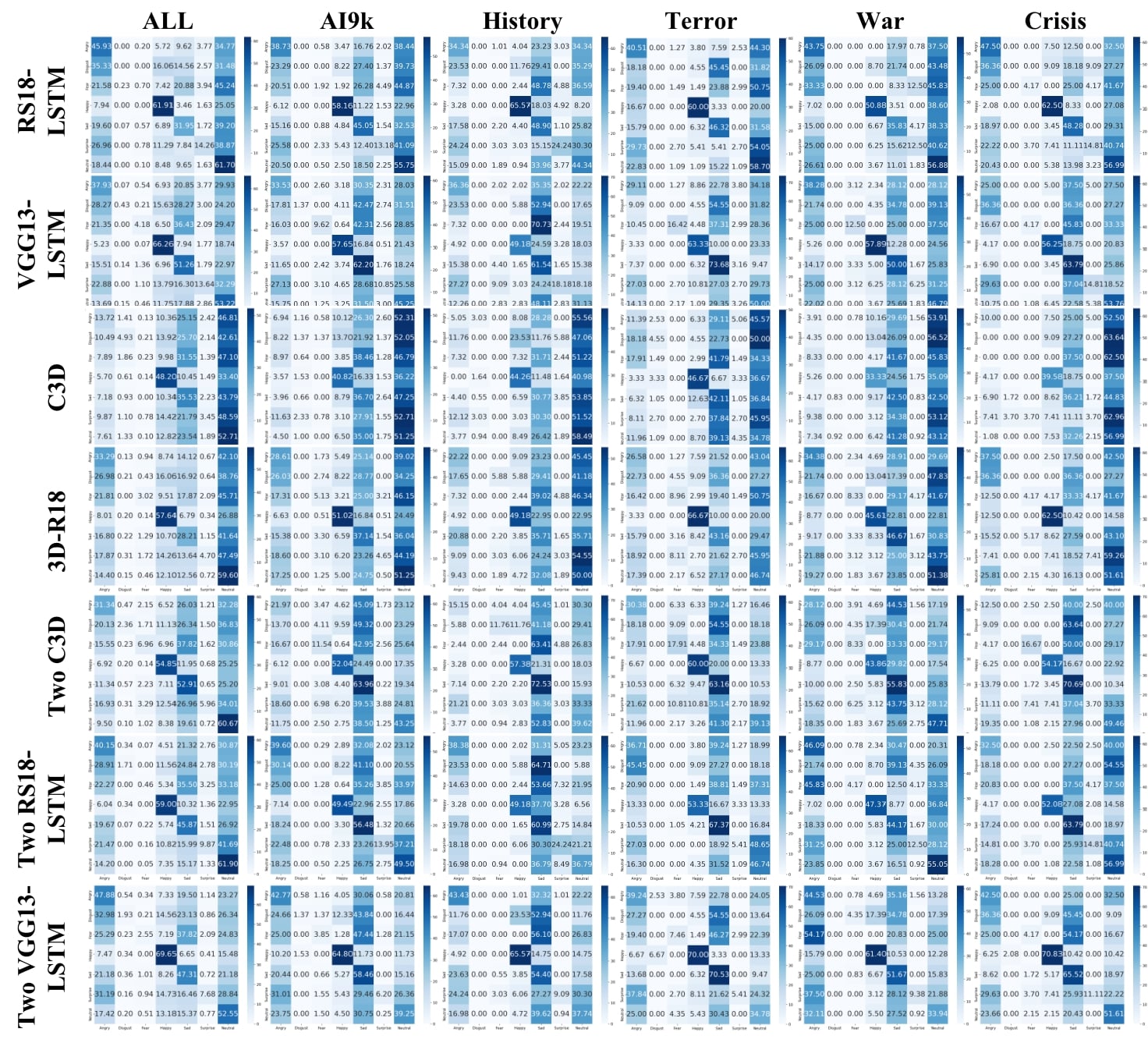}
   \vspace{-0.6cm}
   \caption{Confusion matrices of different methods training on all scenes and then testing on all scenes, AI9k, and its sub-scenes.}
   \label{fig:AI9k}
\end{figure*}

Table \ref{tab:ai_9k} shows comparison of four kinds of baseline architectures training on all scenes and then testing on all scenes, AI9k, and its sub-scenes. The confusion matrices of this experiment is also provided in Figure \ref{fig:AI9k}. And Table \ref{tab:ai_9k_4_to_27} shows the experiment result of intra-scenario performance consistency and inter-scenario invariance via RS50-LSTM training on four scenarios and then testing on AI9k, and its sub-scenes. 

\subsection{Conculsion}
According to above results of four kinds of baseline architectures training on all scenes and then testing on all scenes, four scenarios, and their scenes, we figure out two conclusions: 1) C3D~\cite{W26_tran2015learning} shows the worst results, and VGG13-LSTM achieves the best performance in one-stream network. We consider that C3D~\cite{W26_tran2015learning} fails to capture the temporal information of the limited frames (only 8 frames), but the LSTM can model the global information. 2) Our designed two-stream networks can further improve the performance because the scene context plays an important role in DFER and provides supplemtary information for face-only DFER. In different scenes of our buit FERV39k, confusion matrices of different methods demonstrate that most methods perform best on Happy and perform well on Angry, Sad and Neutral, but are confused in Disgust, Fear, and Surprise due to the limited data and large-scale intensity variations. For cross-domain experiments, the result shows the cross-domain performance of a method is directly related to the feature consistency and intensity of an expression in a scene. For example, WIS9k is designed as a scenario with high similarity and obvious appearance of expressions and the experiment result shows an ideal performance and smaller best-worst difference of each scene among 4 scenarios. 






\begin{table*}[ht]
\centering
\small
\setlength\tabcolsep{0.0085\linewidth}

\begin{tabular}{lcccccccc}
\hline
Method     & All         & DL11k       & Conflict    & School      & DailyLife   & Argue       & Medicine    & Social      \\ \hline
RS18        & 39.33/30.30 & 39.75/31.36 & 39.52/34.65 & 35.80/33.80 & 41.40/31.13 & 44.09/30.81 & 36.36/29.02 & 39.74/33.26 \\
RS50        & 30.57/22.47 & 30.46/21.52 & 31.14/19.23 & 27.41/22.60 & 31.00/19.37 & 36.96/24.07 & 24.48/20.21 & 27.51/25.05 \\
VGG13      & 41.02/31.19 & 40.40/31.59 & 36.53/31.32 & 37.28/34.96 & 39.07/28.63 & 42.40/30.36 & 32.17/23.82 & 48.03/35.50 \\
VGG16      & 41.66/32.01 & 41.81/32.59 & 41.12/34.28 & 38.27/36.03 & 41.19/28.73 & 45.97/29.75 & 31.47/25.16 & 43.23/34.77 \\ \hline
RS18-LSTM   & 42.59/30.92 & 43.34/32.24 & 40.32/30.75 & 37.28/36.31 & 41.61/29.11 & 48.78/30.47 & 41.26/30.32 & 42.36/31.47 \\
RS50-LSTM   & 40.75/32.12 & 40.93/32.91 & 39.92/33.79 & 34.32/32.58 & 41.61/28.00 & 49.53/35.30 & 32.87/27.18 & 42.79/35.70 \\
VGG13-LSTM & 43.37/32.41 & 42.29/32.46 & 43.91/35.84 & 35.31/34.39 & 46.07/31.50 & 46.15/31.31 & 40.56/29.93 & 43.67/34.64 \\
VGG16-LSTM & 41.70/30.93 & 42.99/32.32 & 41.92/33.32 & 37.53/35.84 & 44.37/30.58 & 47.28/31.40 & 40.56/31.08 & 49.34/36.83 \\ \hline
C3D~\cite{W26_tran2015learning}        & 31.69/22.68 & 26.95/21.02 & 21.96/19.35 & 24.94/23.92 & 26.96/18.35 & 31.52/23.00 & 30.77/21.90 & 34.50/24.34 \\
P3D~\cite{W47_qiu2017learning}        & 33.39/23.20 & 32.95/23.80 & 33.53/24.61 & 24.44/23.03 & 32.70/21.39 & 37.34/23.69 & 28.67/21.04 & 34.50/25.05 \\
I3D~\cite{W22_carreira2017quo}        & 38.78/30.17 & 38.56/29.25 & 34.93/26.11 & 36.30/32.55 & 39.70/26.09 & 43.34/28.84 & 34.27/26.52 & 37.55/32.05 \\
3D-RS18~\cite{W45_tran2018closer}    & 37.57/26.67 & 37.69/27.47 & 35.13/25.75 & 32.10/30.63 & 35.67/24.95 & 43.71/28.82 & 27.97/21.50 & 41.48/29.83 \\ \hline
Two C3D    & 41.77/30.72 & 41.45/31.37 & 43.11/35.10 & 33.33/31.64 & 35.46/23.26 & 48.22/31.37 & 33.57/23.14 & 47.16/32.22 \\
Two I3D    & 41.30/31.01 & 41.02/31.55 & 38.92/32.32 & 37.78/33.20 & 40.76/28.93 & 44.84/30.00 & 36.36/23.79 & 44.98/30.94 \\
Two 3D-RS18 & 42.28/30.55 & 42.77/32.72 & 44.31/36.31 & 32.84/30.41 & 39.28/28.41 & 48.41/32.56 & 36.36/24.68 & 49.34/31.62 \\
Two RS18-LSTM   & 43.20/31.28 & 42.20/31.66 & 41.72/31.74 & 36.30/34.63 & 40.55/27.09 & 48.97/32.13 & 37.76/26.91 & 47.60/35.60 \\
Two VGG13-LSTM & 44.54/32.79 & 44.65/32.96 & 43.71/32.61 & 38.52/35.49 & 46.92/31.55 & 50.09/32.30 & 37.76/30.28 & 48.03/36.43 \\ \hline
Average    & 39.58/29.34 & 39.27/29.80 & 38.11/30.24 & 33.90/31.77 & 38.98/26.75 & 44.43/29.60 & 33.99/25.45 & 42.25/31.97 \\ \hline
\end{tabular}
\caption{Comparison of four kinds of baseline architectures training on all scenes and then testing on all scenes, DL11k, and its sub-scenes.}
\label{tab:dl_11k}
\end{table*}

\begin{table*}[ht]
\centering
\small
\setlength\tabcolsep{0.019\linewidth}
\begin{tabular}{lcccccccc}
\hline
Source           & DL11k       & Conflict    & School      & DailyLife   & Argue       & Medicine    & Social      \\ \hline
DL11k      & 38.52/27.72 & 36.53/28.90 & 34.32/29.80 & 37.58/23.58 & 44.84/27.71 & 30.77/25.55 & 37.12/27.02 \\
WIS9k    & 28.00/20.74 & 22.36/17.55 & 20.25/21.39 & 27.60/18.43 & 32.46/20.42 & 29.37/21.13 & 39.30/27.49 \\
SIA10k     & 28.05/21.61 & 20.56/18.30 & 24.69/24.66 & 32.27/22.14 & 32.27/22.40 & 29.37/19.57 & 36.24/24.55 \\ 
AI9k      & 26.38/19.95 & 22.55/17.45 & 23.95/22.52 & 26.54/17.86 & 26.08/19.43 & 23.78/19.72 & 28.82/27.80 \\
\hline
\end{tabular}
\caption{Experiment results of intra-scenario performance consistency (highlighted in blue bar) and inter-scenario invariance via RS50-LSTM training on four scenarios and then testing on DL11k, and its sub-scenes.}
\label{tab:dl_11k_4_to_27}
\end{table*}

\begin{table*}[ht]
\centering
\small
\setlength\tabcolsep{0.0075\linewidth}

\begin{tabular}{lcccccccc}
\hline
Method     & All         & WIS9k       & Action      & ScholarReport & Speech      & Liveshow    & ElegantArt  & Talkshow    \\ \hline
RS18        & 39.33/30.30 & 40.50/28.67 & 50.61/34.11 & 40.25/27.59   & 43.09/31.19 & 37.72/26.82 & 33.33/25.53 & 38.57/25.47 \\
RS50        & 30.57/22.47 & 32.52/23.50 & 37.80/27.77 & 35.31/24.30   & 30.89/23.22 & 28.51/23.13 & 31.35/21.51 & 28.86/20.14 \\
VGG13      & 41.02/31.19 & 43.04/30.23 & 54.88/34.08 & 47.41/33.11   & 38.48/28.07 & 44.74/30.40 & 36.51/28.53 & 40.57/26.25 \\
VGG16      & 41.66/32.01 & 42.93/30.77 & 53.05/33.57 & 43.21/29.00   & 38.75/29.22 & 46.05/33.65 & 40.87/32.16 & 39.43/24.96 \\ \hline
RS18-LSTM   & 42.59/30.92 & 44.12/29.59 & 51.83/31.02 & 44.94/27.73   & 39.57/27.95 & 46.93/31.59 & 37.70/30.48 & 44.57/27.23 \\
RS50-LSTM   & 40.75/32.12 & 41.74/30.70 & 56.10/36.76 & 43.70/30.88   & 39.57/29.34 & 40.35/30.40 & 38.89/32.57 & 40.00/27.42 \\
VGG13-LSTM & 43.37/32.41 & 44.23/30.81 & 53.05/33.84 & 45.19/30.82   & 40.92/30.13 & 45.61/31.28 & 41.27/31.19 & 44.29/29.17 \\
VGG16-LSTM & 41.70/30.93 & 41.63/28.42 & 46.95/27.49 & 45.19/30.54   & 42.82/30.86 & 36.84/25.76 & 39.68/30.48 & 41.14/26.39 \\ \hline
C3D~\cite{W26_tran2015learning}        & 31.69/22.68 & 30.15/19.94 & 35.98/20.38 & 26.91/16.95   & 31.71/21.92 & 28.51/22.55 & 28.97/21.60 & 36.57/23.25 \\
P3D~\cite{W47_qiu2017learning}       & 33.39/23.20 & 34.95/22.40 & 39.63/24.87 & 36.30/22.10   & 36.31/23.94 & 34.65/23.13 & 27.38/19.42 & 31.43/18.32 \\
I3D~\cite{W22_carreira2017quo}         & 38.78/30.17 & 38.52/29.11 & 46.34/31.67 & 46.17/35.70   & 34.42/25.81 & 37.72/30.57 & 34.92/25.59 & 26.29/18.27 \\
3D-RS18~\cite{W45_tran2018closer}     & 37.57/26.67 & 38.40/24.85 & 48.78/28.40 & 38.52/23.48   & 39.30/27.40 & 39.04/25.12 & 32.54/24.44 & 36.29/21.86 \\ \hline
Two C3D    & 41.77/30.72 & 43.44/29.77 & 54.27/35.23 & 43.70/30.48   & 41.73/30.17 & 41.23/25.74 & 42.06/28.16 & 42.00/27.89 \\
Two I3D    & 41.30/31.01 & 42.31/30.14 & 57.93/36.47 & 46.67/31.42   & 40.92/30.57 & 38.16/25.91 & 37.30/28.55 & 39.43/28.37 \\
Two 3D-RS18 & 42.28/30.55 & 44.12/29.63 & 54.88/30.61 & 46.91/29.62   & 42.28/30.17 & 41.67/28.50 & 36.11/27.20 & 38.57/24.66 \\
Two RS18-LSTM   & 43.20/31.28 & 44.91/30.37 & 57.32/34.12 & 47.16/31.88 & 41.46/30.12 & 44.74/26.55 & 37.70/26.76 & 43.43/27.52 \\
Two VGG13-LSTM & 44.54/32.79 & 45.25/31.45 & 57.93/38.26 & 47.90/32.43 & 40.11/29.40 & 48.25/33.02 & 43.65/31.93 & 45.14/28.30 \\ \hline
Average    & 39.58/29.34 & 40.61/28.11 & 50.23/31.51 & 42.68/28.69   & 38.91/28.15 & 39.79/27.65 & 36.15/27.23 & 38.39/24.75 \\ \hline
\end{tabular}
\caption{Comparison of four kinds of baseline architectures training on all scenes and then testing on all scenes, WIS9k, and its sub-scenes.}
\label{tab:wis_9k}
\end{table*}

\begin{table*}[ht]
\centering
\small
\setlength\tabcolsep{0.019\linewidth}

\begin{tabular}{lcccccccc}
\hline
Source         & WIS9k       & Action      & ScholarReport & Speech      & Liveshow    & ElegantArt  & Talkshow    \\ \hline
DL11k      & 30.71/20.24 & 46.34/31.23 & 26.42/16.74   & 31.98/23.59 & 28.95/20.16 & 28.57/21.66 & 30.00/19.10 \\
WIS9k     & 40.72/26.88 & 51.22/29.88 & 42.96/28.01   & 39.02/25.83 & 39.47/23.02 & 36.11/24.98 & 38.86/24.77 \\
SIA10k    & 30.94/19.79 & 37.20/20.95 & 27.90/16.87   & 34.42/24.20 & 26.32/16.58 & 29.37/20.65 & 34.29/20.87 \\ 
AI9k      & 23.25/17.25 & 20.73/18.17 & 20.00/14.75   & 27.37/21.01 & 18.42/12.36 & 25.79/18.49 & 22.57/16.90 \\
\hline 
\end{tabular}
\caption{Experiment results of intra-scenario performance consistency (highlighted in blue bar) and inter-scenario invariance via RS50-LSTM training on four scenarios and testing on WIS9k, and its sub-scenes.}
\label{tab:wis_10k_4_to_27}
\end{table*}

\begin{table*}[ht]
\centering
\small
\setlength\tabcolsep{0.0080\linewidth}

\begin{tabular}{lcccccccc}
\hline
Method     & All         & SIA10k      & Business    & Experiment  & OfficialEvent & Crime       & Interview   & Contest     \\ \hline
RS18        & 39.33/30.30 & 42.31/30.02 & 38.83/28.01 & 49.56/26.70 & 37.98/25.48   & 36.34/28.63 & 45.75/29.18 & 48.24/33.37 \\
RS50        & 30.57/22.47 & 30.56/22.68 & 22.07/17.53 & 31.86/16.89 & 34.49/20.96   & 23.08/17.26 & 33.25/21.72 & 37.06/27.55 \\
VGG13      & 41.02/31.19 & 43.44/29.99 & 39.39/27.83 & 49.56/26.52 & 38.33/26.04   & 36.87/27.95 & 44.34/28.83 & 47.62/32.39 \\
VGG16      & 41.66/32.01 & 42.31/29.58 & 37.43/26.94 & 52.21/33.12 & 41.46/26.09   & 34.75/28.20 & 47.17/30.77 & 48.03/32.92 \\ \hline
R18-LSTM   & 42.59/30.92 & 42.85/28.78 & 41.34/29.45 & 48.67/25.42 & 37.98/22.40   & 40.32/31.14 & 45.52/28.01 & 50.10/33.79 \\
R50-LSTM   & 40.75/32.12 & 42.16/30.39 & 37.99/28.47 & 47.79/33.17 & 39.37/26.84   & 36.07/29.63 & 43.87/30.02 & 48.24/34.32 \\
VGG13-LSTM & 43.37/32.41 & 45.00/31.45 & 43.02/31.22 & 57.52/36.17 & 37.63/23.05   & 40.05/31.33 & 47.17/30.07 & 49.90/33.66 \\
VGG16-LSTM & 41.70/30.93 & 43.83/29.83 & 40.22/29.55 & 53.10/30.03 & 40.77/24.54   & 37.40/29.00 & 46.23/27.39 & 48.65/34.15 \\ \hline
C3D~\cite{W26_tran2015learning}       & 31.69/22.68 & 42.70/29.22 & 40.78/28.44 & 54.87/22.87 & 35.89/21.74   & 36.07/25.16 & 43.16/26.35 & 46.58/32.44 \\
P3D~\cite{W47_qiu2017learning}         & 33.39/23.20 & 36.73/23.66 & 33.24/21.28 & 42.48/21.98 & 32.75/19.02   & 32.10/21.69 & 39.62/21.94 & 40.58/26.69 \\
I3D~\cite{W22_carreira2017quo}        & 38.78/30.17 & 40.55/31.07 & 33.52/26.20 & 53.10/31.56 & 38.68/28.47   & 38.46/29.99 & 41.51/27.87 & 45.55/35.43 \\
3D-R18~\cite{W45_tran2018closer}      & 37.57/26.67 & 40.40/26.08 & 35.75/23.46 & 54.87/32.50 & 37.63/21.68   & 33.69/23.53 & 42.69/22.70 & 44.10/28.32 \\ \hline
Two C3D    & 41.77/30.72 & 44.71/30.15 & 46.09/35.27 & 63.72/37.55 & 35.89/22.51   & 38.99/27.51 & 46.23/28.45 & 48.03/32.31 \\
Two I3D    & 41.30/31.01 & 43.63/31.20 & 38.83/30.19 & 54.87/26.96 & 39.72/25.10   & 40.05/27.89 & 44.81/29.96 & 48.03/33.28 \\
Two 3D-R18 & 42.28/30.55 & 42.95/27.83 & 38.83/26.57 & 62.83/33.41 & 33.80/19.25   & 38.99/26.28 & 45.52/24.71 & 48.45/33.16 \\
Two RS18-LSTM   & 43.20/31.28 & 46.33/31.09 & 44.69/31.57 & 57.52/24.56 & 36.93/21.72 & 38.20/28.59 & 47.41/28.50 & 53.00/33.93 \\
Two VGG13-LSTM & 44.54/32.79 & 46.57/31.88 & 44.69/32.33 & 53.98/31.66 & 43.90/26.60 & 38.46/30.10 & 46.70/28.35 & 52.80/35.32 \\ \hline
Average    & 39.58/29.34 & 42.04/28.94 & 38.25/27.69 & 52.06/28.57 & 37.41/23.30   & 36.28/27.15 & 44.19/27.22 & 47.33/32.39 \\ \hline
           &             &             &             &             &               &             &             &            
\end{tabular}
\caption{Comparison of four kinds of baseline architectures training on all scenes and then testing on all scenes, SIA10k, and its sub-scenes.}
\label{tab:sia_10k}
\end{table*}

\begin{table*}[ht]
\centering
\small
\setlength\tabcolsep{0.019\linewidth}

\begin{tabular}{lcccccccc}
\hline
Source    & SIA10k      & Business    & Experiment  & OfficialEvent & Crime       & Interview   & Contest     \\ \hline
DL11k      & 30.85/21.86 & 23.46/18.93 & 38.05/26.28 & 31.01/19.72   & 26.53/21.73 & 33.73/20.12 & 34.78/24.44 \\
WIS9k     & 32.32/20.59 & 30.17/21.69 & 38.94/21.20 & 29.27/17.46   & 28.91/21.35 & 33.73/17.19 & 33.75/21.78 \\
SIA10k  & 39.96/25.22 & 36.87/23.59 & 63.72/42.51 & 36.24/21.66   & 33.95/22.52 & 41.27/22.95 & 44.93/28.35 \\ 
AI9k      & 24.53/18.79 & 16.76/13.15 & 31.86/28.26 & 23.00/15.02   & 25.20/20.87 & 25.94/17.42 & 27.33/19.83 \\
\hline
\end{tabular}
\caption{Experiment results of intra-scenario performance consistency (highlighted in blue bar) and inter-scenario invariance via RS50-LSTM training on four scenarios and testing on SIA10k, and its sub-scenes.}
\label{tab:sia_10k_4_to_27}
\end{table*}

\begin{table*}[ht]
\setlength\tabcolsep{0.0185\linewidth}
\centering
\begin{tabular}{lcccccc}
\hline
Method     & All         & AI9k        & History     & Terror      & War         & Crisis      \\ \hline
RS18        & 39.33/30.30 & 33.90/27.20 & 31.17/23.72 & 31.28/26.69 & 35.09/28.30 & 36.88/29.21 \\
RS50        & 30.57/22.47 & 30.14/19.94 & 35.44/21.05 & 26.54/19.67 & 30.83/19.19 & 24.25/20.11 \\
VGG13      & 41.02/31.19 & 38.86/29.94 & 36.92/25.75 & 36.73/31.48 & 37.73/29.22 & 42.52/31.65 \\
VGG16      & 41.66/32.01 & 39.60/31.46 & 37.11/26.81 & 39.57/34.21 & 41.38/32.80 & 40.86/32.95 \\ \hline
R18-LSTM   & 42.59/30.92 & 39.66/30.40 & 40.45/31.06 & 35.55/29.96 & 39.35/28.55 & 44.85/33.46 \\
R50-LSTM   & 40.75/32.12 & 38.01/31.16 & 40.07/30.81 & 36.26/32.46 & 39.15/31.26 & 39.87/31.87 \\
VGG13-LSTM & 43.37/32.41 & 41.20/31.49 & 40.26/28.06 & 40.28/33.61 & 40.16/30.24 & 42.86/31.11 \\
VGG16-LSTM & 41.70/30.93 & 37.04/29.39 & 38.03/28.82 & 36.26/32.83 & 36.92/27.63 & 41.53/33.59 \\ \hline
C3D~\cite{W26_tran2015learning}       & 31.69/22.68 & 27.29/19.80 & 27.83/19.80 & 22.99/20.31 & 24.75/17.55 & 32.56/20.93 \\
P3D~\cite{W47_qiu2017learning}       & 33.39/23.20 & 31.34/21.52 & 32.65/21.23 & 27.96/22.71 & 31.03/21.10 & 34.55/21.61 \\
I3D~\cite{W22_carreira2017quo}        & 38.78/30.17 & 37.44/28.15 & 40.07/29.02 & 33.89/29.10 & 37.53/26.32 & 36.54/28.81 \\
3D-R18~\cite{W45_tran2018closer}      & 37.57/26.67 & 33.45/25.40 & 31.73/22.88 & 31.28/27.83 & 37.53/27.07 & 37.21/27.25 \\ \hline
Two C3D    & 41.77/30.72 & 37.89/28.09 & 41.93/27.16 & 35.78/30.47 & 37.12/26.71 & 40.86/29.60 \\
Two I3D    & 41.30/31.01 & 38.75/28.53 & 38.78/26.49 & 36.02/29.19 & 38.95/29.20 & 38.87/28.01 \\
Two 3D-R18 & 42.28/30.55 & 38.46/28.54 & 40.45/28.11 & 35.07/28.73 & 36.71/26.55 & 42.19/29.58 \\
Two RS18-LSTM   & 43.20/31.28 & 40.40/30.04 & 41.93/29.94 & 36.49/29.94 & 41.38/29.91 & 43.85/31.45 \\
Two VGG13-LSTM & 44.54/32.79 & 40.63/30.96 & 41.74/30.03 & 37.44/32.49 & 39.35/28.70 & 46.84/35.11 \\ \hline
Average    & 39.58/29.34 & 36.55/27.61 & 37.41/26.45 & 33.75/28.70 & 36.53/26.93 & 39.12/29.12 \\ \hline
\end{tabular}
\caption{Comparison of four kinds of baseline architectures training on all scenes and then testing on all scenes, AI9k, and its sub-scenes.}
\label{tab:ai_9k}
\end{table*}

\begin{table*}[ht]
\centering
\setlength\tabcolsep{0.035\linewidth}

\begin{tabular}{lcccccc}
\hline
Source           & AI9k        & History     & Terror      & War         & Crisis      \\ \hline
DL11k      & 25.64/19.58 & 27.09/21.19 & 23.70/20.43 & 24.34/16.88 & 24.58/18.86 \\
WIS9k     & 23.65/18.43 & 25.23/20.01 & 21.80/20.42 & 24.34/17.10 & 28.24/19.57 \\
SIA10k      & 27.92/20.30 & 29.68/21.05 & 24.64/19.85 & 27.38/20.03 & 30.23/19.32 \\ 
AI9k       & 31.68/24.04 & 31.73/21.07 & 28.67/24.87 & 34.48/24.25 & 31.56/24.63 \\
\hline    
\end{tabular}
\caption{Experiment results of intra-scenario performance consistency (highlighted in blue bar) and inter-scenario invariance via RS50-LSTM training on four scenarios and testing on AI9k, and its sub-scenes.}
\label{tab:ai_9k_4_to_27}
\end{table*}

\end{document}